\documentclass{article}

\usepackage{etoc}               
\usepackage[inline]{enumitem}
\usepackage{multirow}       
\usepackage{subcaption}     
\usepackage{wrapfig}        
\usepackage{amsmath}
\usepackage{amssymb}
\usepackage{mathtools}
\usepackage{amsthm}
\usepackage{algorithm}
\usepackage{algorithmic}

\theoremstyle{plain}
\newtheorem{theorem}{Theorem}[section]
\newtheorem{proposition}[theorem]{Proposition}
\newtheorem{lemma}[theorem]{Lemma}
\newtheorem{corollary}[theorem]{Corollary}
\theoremstyle{definition}

\theoremstyle{remark}

\DeclareMathOperator*{\argmin}{arg\,min}

\PassOptionsToPackage{numbers, compress}{natbib}
 \usepackage[preprint]{neurips_2026}

\usepackage[utf8]{inputenc} 
\usepackage[T1]{fontenc}    
\usepackage[dvipsnames]{xcolor}
\usepackage[pagebackref=true, colorlinks=true, linkcolor=Blue, urlcolor=Blue, citecolor=Blue]{hyperref}       
\usepackage{url}            
\usepackage{booktabs}       
\usepackage{amsfonts}       
\usepackage{nicefrac}       
\usepackage{microtype}      

\title{Efficient compression of neural networks and datasets}

\author{%
  Lukas Silvester Barth\thanks{Equal contribution. Contributions are detailed in Appendix~\ref{app:contributions}.} \\
  \texttt{lukas.barth@mis.mpg.de} \\
  \And
  Paulo von Petersenn\footnotemark[1] \\
  \texttt{vonpetersenn@mis.mpg.de} \\
  \AND
  \normalfont
  Max Planck Institute for Mathematics in the Sciences, Leipzig, Germany
}

\begin{document}

\maketitle

\begin{abstract}
  Compression and generalization are fundamentally related through Solomonoff induction and the minimum description length principle (MDL), which predict that simpler models generalize better when data arises from low-complexity distributions. 
  In this article, we combine insights from algorithmic information theory and techniques from neural network pruning to improve model generalization by identifying the most effective data compression method.
  Since exact MDL optimization is intractable, we cast it as $\ell_0$ regularized learning and explain why parameter sparsity provides an effective computable approximation of model description length.
  To identify the best practical approach, we systematically compare and refine complementary sparse optimization methods. 
  In particular, we improve probabilistic pruning through a procedure that does not require Monte Carlo sampling and refine smooth $\ell_0$ approximations with a binary search routine that reduces hyperparameter complexity.
  Across convolutional networks and transformers evaluated on image and text datasets, our refined methods improve upon their predecessors, achieve substantial model compression with minimal accuracy loss, and yield short data description lengths. 
  Finally, we use these methods in a controlled teacher-student setting to empirically verify the prediction of Solomonoff induction that compressed models learn more sample-efficiently and generalize better.
\end{abstract}

\vspace{0.5cm}
\makeatletter
\let\l@part\l@section 
\makeatother
\addtocontents{toc}{\protect\setlength{\parskip}{0pt}} 
\setcounter{tocdepth}{1}
\tableofcontents
\newpage


\section{Introduction}
\label{Introduction}

A central challenge in machine learning is to extract the most from available data by capturing the underlying structure of a problem. 
One key insight of the transformer architecture \cite{vaswani2017attention} is that scaling in model size, data, and compute yields largely monotonic performance gains,
which has driven recent advancements in large language models, image generation, and video diffusion \cite{rae2021scaling, hoffmannTrainingComputeOptimalLarge, touvron2023llama, esser2024scaling, imagenteamgoogle2024imagen3}.
However, scaling alone leaves open how to best exploit a fixed parameter budget.
Many extensions of the original transformer architecture exploit architectural forms of parameter sharing or sparsity \cite{jiang2024mixtralexperts, deepseekai2024deepSeekV3technicalReport}.
While effective, such approaches are architecture- and task-specific, relying on manual design choices rather than general learning objectives.
This motivates the study of learning principles that yield compact yet well-performing models without architecture-specific assumptions.

One such general, theoretically principled learning framework is compression. 
Since compression eliminates redundancy by finding rules under which the data could have been generated, it is closely tied to generative AI, model identification, and regularization.
Formal notions of information and compressibility trace back to Shannon's source coding theorem \cite{shannon1948mathematical} and Kolmogorov complexity \cite{kolmogorov1963tables}.
Closely related, Solomonoff induction \cite{solomonoff1964formal} connects compression to probabilistic inference and yields strong posterior consistency guarantees \cite{solomonoff1978complexity}.
Building on similar ideas, the minimum description length (MDL) principle casts learning itself as a problem of optimal compression \cite{rissanen1978modeling,grunwald2007minimum}.
MDL learning selects the model minimizing the combined description length of the model itself and the data it explains.
Under standard assumptions, this criterion offers formal guarantees that the selected model generalizes well \cite{barron2002minimum,poland2004convergence}.
In the area of signal processing, compressive sensing \cite{OrthogonalMatchinPursuit1993,tibshirani1996regression,blumensath2008iterative, foucart_mathematical_2013} studies sparse recovery as a means of inferring low-complexity structure from limited data.
Furthermore, algorithmic information theory has been related to sequential decision theory and multi-agent learning \cite{hutter2005universal,meulemans2025embedded}.
When viewed as a learning principle, compression is not merely a post-hoc operation to increase computational efficiency but induces a bias toward simpler explanations that can improve generalization and sample efficiency. 

In practice, the smallest description length is uncomputable \cite{shen2017kolmogorov} and must be approximated.
For parametric models such as neural networks, a natural approximation is to penalize the number of active parameters.
This leads to $\ell_0$ regularized objectives, which directly control the model size rather than weight magnitude and provide a tractable surrogate for description length minimization.
Modern deep learning brings new practical relevance to these historical ideas because expressive probabilistic models trained by backpropagation can now compress highly complex data distributions \cite{DeletangICLR2024}.

\textbf{Our aim} is to bridge the gap between algorithmic information theory and practical compression methods in order to identify the most efficient pruning procedures for improving model generalization.

\textbf{Contributions}
\begin{enumerate}
    \item We provide the theoretical link between description length minimization and $\ell_0$ regularization as a natural basic surrogate for model complexity. 
    To verify the theoretical predictions, we introduce a teacher-student setting that allows to conduct controlled experiments, which confirm that compression indeed improves generalization and sample efficiency when the data-generating process is simple. 
    \label{contribution:1}
    \item We contribute to research on how non-differentiable and NP-hard $\ell_0$ regularized learning objectives can be optimized with neural networks and backpropagation. To do so, 
    we improve probabilistic pruning through a procedure that
    does not require Monte Carlo sampling and refine smooth approximations with
    a binary search routine that reduces hyperparameter complexity.
    We also provide auxiliary techniques such as layerwise pruning and random gradient pruning. 
    \label{contribution:2}
    \item We empirically benchmark compression-regularized training across diverse architectures and datasets, including transformer language models and image classifiers. 
    Our methods effectively outperform their predecessors and reduce model size while maintaining or improving test performance. 
    This shows that compression can act as a general inference principle. 
    \label{contribution:3}
    \item We publish detailed open source code for our efficient compression procedures in both Python (PyTorch) and Julia (Lux) at \textbf{\href{https://github.com/L0-and-behold/efficient-compression}{github.com/L0-and-behold/efficient-compression}}.
\end{enumerate}

\section{Related work}
\label{sec:relatedWork}

We were inspired by the idea that \textbf{compression, description length minimization and Solomonoff induction} can take the role of general learning principles that guarantee the statistical consistency of the learning process \cite{rissanen1978modeling,solomonoff1978complexity,barron2002minimum,hutter2005universal,grunwald2007minimum}. However, these works do not directly connect these ideas to modern machine learning methods because it is hard to bring the description length objective into a form that is amenable to backpropagation. The core problem is that the model size in the objective is non-differentiable.
One simple approach to circumvent this problem, called \textbf{offline compression}, is to fix the model size during training and compress the data via arithmetic coding \cite{pasco1976source,RissanenArithmeticCoding} after training \cite{deletang2023language}. 
Another approach, called \textbf{online compression} or prequential MDL \cite{philip1999prequential,bornschein2023sequential,JackRaeVideo} views training as sequential communication: The sender transmits an untrained model, then iteratively compresses and sends each data batch, which the receiver can decode with the model trained on previously received data.
In this formulation, the area under the negative log-likelihood loss curve directly measures how many bits are needed to transmit the dataset \cite{JackRaeVideo}. 
However, both approaches do not adapt the model size during training. 
In contrast, we include an approximation of the model size in the learning objective and show that this yields better compression results (Sec.~\ref{sec:transformerExperiments} and App.~\ref{app:llama_analysis}).

The upper bound we chose is the bitlength per parameter multiplied with the $\ell_0$ norm of the network. This choice is most canonical in the sense that it does not require an encoding distribution, as we explain in more detail in Sec.~\ref{sec:tractableApproximation}, and relates our work to \textbf{neural network pruning} on which there is a large body of literature (see citations in Tab.~\ref{tab:cifar-vgg} and \ref{tab:imagenet-resnet-50}, as well as in Tab.~\ref{tab:mnist-lenet300} in App.~\ref{sec:classifier-benchmark-tables}). However, these methods usually treat model compression as a post-hoc engineering task 
as opposed to a general learning principle. We provide the link to MDL (Sec.~\ref{theory}) and verify theoretical predictions in controlled experiments (Sec.~\ref{sec:stExperiments}).
Additionally, we implement and open-source $\ell_0$ regularization algorithms that outperform previous compression and pruning baselines (Sec.~\ref{sec:ImageClassifierExperiments}).

\textbf{$\ell_0$ regularized optimization} is itself an NP-hard problem \cite{foucart_mathematical_2013} and requires differentiable approximations to become amenable to backpropagation. Probabilistic reformulations that admit gradient descent were developed in \cite{yin2020probabilistic,dong2020disarm,kunes2023gradient} and \cite{louizos2017learning}. However, they require Monte Carlo sampling. We improve their efficiency with a formulation that does not require this (Sec.~\ref{sec:probabilistic-reformulations}). Another line of work replaces the $\ell_0$ norm with a piecewise smooth function. We build on \cite{gu2009l0,oliveira2024compression} who employ an exponential decay term and on (relaxed) $\ell_1$ regularization known from compressive sensing \cite{tibshirani1996regression,meinshausen2007relaxed,hastie2017extended} but also employed in nonlinear models \cite{sahoo2018learning}. We improve these methods with several auxiliary techniques (Sec.~\ref{sec:supportingMethods}) and identify unnecessary parameters through ablation studies (App.~\ref{app_ablation_studies}).
\section{Theory}
\label{theory}

\subsection{Preliminaries}
\label{sec:preliminaries}
All notions of compression can be subsumed under a Bayesian viewpoint when adopting priors that encode model complexity. 
We present this viewpoint, starting from Kolmogorov complexity \cite{kolmogorov1963tables}.

Let $U$ be a universal reference machine (e.g.~a universal Turing machine), $x$ a finite string, $\mathcal{P}$ the set of terminating programs (encoded as strings), and $\ell(p)$ the bit length of $p \in \mathcal{P}$. 
The Kolmogorov complexity $K_U$ is then 
\begin{equation}
        K_U(x) := \min_{p\in\mathcal{P}}\{ ~\ell(p)~|~U(p)=x~\}.
    \label{eq:KolmComp}
\end{equation}
$K_U$ measures the length of (one of) the shortest program(s) generating $x$. 
For compression, we focus on approximating (one of) the shortest program(s) $p^*(x)$ generating $x$. 

Arithmetic coding \cite{pasco1976source,RissanenArithmeticCoding} allows to use any probability distribution $\mu$ to encode any finite sequence $x$ with $-\log_2(\mu(x))$ bits (up to a small constant). 
By Shannon's source coding theorem \cite{shannon1948mathematical}, this is optimal on average (w.r.t.~$\mu$) and the average equals the entropy of $\mu$. 
However, when accounting for the description of $\mu$ itself, the total description length $L_\mu(x)$ of $x$ becomes
\begin{equation}
    L_\mu(x) = \ell(\mu)-\log_2(\mu(x)),
    \label{eq:descriptionLength}
\end{equation}
where $\ell$ measures the bits needed to describe $\mu$ as a program of $U$.
In machine learning, $\ell(\mu)$ matters when model sizes exceed data sizes. 
Eq.~\eqref{eq:descriptionLength} is the fundamental object in the theory of (minimum) description length (MDL) \cite{rissanen1978modeling,grunwald2007minimum}. 
If $\mathcal{M}$ is the space of all computable distributions, then $\min_{\mu\in \mathcal{M}}L_\mu(x)$ and $K_U(x)$ are equivalent up to a constant that does not depend on $x$, see App.~\ref{app:equivalenceOfKolmCompAndDescLength}.

One advantage of Eq.~\eqref{eq:descriptionLength} over Eq.~\eqref{eq:KolmComp} is that it relates description length to probability theory and allows to separate information about the distribution from information about individual samples or noise.
It is useful to rewrite the argmin of Eq.~\eqref{eq:descriptionLength} as an argmax:
\begin{equation}
    \text{argmin}_\mu ~L_\mu(x)
    =\text{argmax}_\mu~ 2^{-\ell(\mu)}\mu(x)
    \label{eq:argmaxformulation}
\end{equation}
If we define a ``complexity prior'' over the distributions $\mu$ by $p(\mu) \propto 2^{-\ell(\mu)}$, then Bayes' rule gives the Bayesian posterior probability 
$p(\mu|x) = p(\mu)p(x|\mu)/p(x)\propto 2^{-\ell(\mu)} \mu(x)$. 
Thus $L_\mu(x)$ (and $K_U(x)$) is the maximum a posterior estimator (MAP) of the posterior $p(\mu|x)$. 

Eq.~\eqref{eq:argmaxformulation} uses a maximum, replacing it by a sum yields the Bayesian predictive prior distribution:
\begin{equation}
    \xi(x) \propto \sum_{\mu \in \mathcal{M}} 2^{-\ell(\mu)} \mu(x) 
    \label{eq:predictivePrior}
\end{equation}
This, in turn, is the fundamental object in the theory of Solomonoff induction \cite{solomonoff1964formal}. Its importance in particular stems from a strong posterior consistency result \cite{solomonoff1978complexity,hutter2005universal}:
\begin{theorem}
    \label{thm:SolomonoffInduction}
    Given any $\mu\in\mathcal{M}$, let $\xi(x)$ be a (semi-) probability distribution that satisfies $\xi(x) \ge w(\mu)\, \mu(x)$ for some fixed $w(\mu)$ and all $x$. Then, for all $n\in\mathbb{N}$,
    \begin{align*}
            \sum_{t=1}^n \mathbb{E}_{x_{< t}\sim \mu}\bigg[\sum_{x_t}(\mu(x_t|x_{<t})-\xi(x_t|x_{<t}))^2\bigg] \le \ln w_\mu^{-1}.
    \end{align*}
\end{theorem}
For the MAP (see Eq.~\eqref{eq:argmaxformulation}), similar consistency theorems can be proven \cite{poland2004convergence} albeit with slower convergence rate, namely with an exponentially larger right-hand-side bound. 
The key implication of these theorems is that complexity priors lead to sample-efficient convergence, if the true data-generating model $\mu$ has low complexity. 

To summarize, while Kolmogorov complexity constitutes the gold standard of compression, the description length (compare Eq.~\eqref{eq:descriptionLength}) serves as an equivalent objective that is easier to approximate. 
The posterior consistency results from Solomonoff induction provide guarantees that compression recovers the true distribution $\mu$ underlying the data. 

\subsection{Computationally tractable \texorpdfstring{$\ell_0$}{L0} approximation}
\label{sec:tractableApproximation}
Due to the Halting problem, the Kolmogorov complexity is uncomputable \cite{shen2017kolmogorov}, but it can be approximated from above. 
While the general space of computable distributions is intractable, we can obtain upper bounds by optimizing over an expressive family of parameterized probability distributions $\mu_\theta$, where $\theta$ denotes a high-dimensional parameterization. 
In this case, the description length becomes $\ell(\mu_\theta)-\log_2 \mu_\theta(x)$ and the term $\ell(\mu_\theta)$ can be expressed as a function of $\theta$. 
To encode parameters $\theta$, an optimal choice according to Shannon's source coding theorem would be to introduce a prior distribution $p$ and to use entropy encoding $\ell(\theta|p) = -\log_2 p(\theta)$. 
However, this prior must itself be encoded, leading to a hierarchy:  $\ell(\theta) = \ell(\theta|p) + \ell(p|p') + \ell(p'|p'') + \cdots$. 
This regress terminates when we reach a base-level prior that requires no further specification---typically a uniform distribution over a discrete set. 
For parameters in $\mathbb{R}^d$, this means quantizing to a finite grid and using a uniform prior over grid points, which yields $\ell(\theta) \propto$ \texttt{(number of parameters $\times$ bits per parameter)}. 
Thus, the hierarchy of increasingly informative priors ultimately grounds out in the $\ell_0$ norm: the combinatorial complexity of specifying which parameters are non-zero or active. 
In this sense, parameter count is the fundamental measure of model complexity as it represents the irreducible description length when all distributional structure is stripped away.

This insight motivates us to study the $\ell_0$ regularization problem as an approximation of Eq.~\eqref{eq:descriptionLength},
$L_\theta(x)  = \alpha \ell_0(\theta) - \log_2 \mu_\theta(x)$,
where $\ell_0(\theta)=|\{~i~|~\theta_i\ne 0~\}|$
is the so-called {\it L-zero} norm. 
This approximation preserves the connection to Solomonoff induction:
Even when replacing the exact description length $\ell$ by $\ell_0$, adopting the corresponding prior $w_\mu \propto 2^{-\ell_0(\mu)}$ in Eq.~\eqref{eq:argmaxformulation} and Eq.~\eqref{eq:predictivePrior} retains the posterior consistency guarantees from Thm.~\ref{thm:SolomonoffInduction}, as well as its MAP (MDL) approximation.

In the next sections, we will derive several methods that find local minima of $L_\theta(x)$, thereby providing approximate solutions to the compression problem. 
These will require the parameterized function family $\mu_\theta$ to be piecewise differentiable, but extend to objectives of the form 
\begin{equation}
        L_\theta(x) = \alpha \ell_0(\theta) - \mathcal{L}(f_\theta, x),
    \label{eq:l0objective}
\end{equation}
where $\mathcal{L}(f_\theta, x)$ is any piecewise differentiable loss function.
If $\mu_\theta$ is computed by a neural network, then sparse matrix representations (COO or LIL) can be used to encode the weights of $\ell_0$ regularized networks efficiently. 

\subsection{Mean square error and teacher student setup}
\label{sec:mseandteacherstudentsetup}
The loss function \eqref{eq:descriptionLength} encodes data description length. It turns out that we can recover the
mean squared error (MSE) loss as special case of Eq.~\eqref{eq:descriptionLength} when restricting $\mathcal{M}$ to conditional Gaussian distributions with mean $m(\cdot)$ and standard deviation $\sigma$, as we explain below. 
We use this insight to derive a refined loss that co-optimizes noise variance (App.~\ref{app:refinedSquareLoss}) and construct teacher-student experiments testing Solomonoff induction's sample efficiency predictions.

Suppose the data-generating process is given by a conditional Gaussian distribution with mean $m$ and variance $\sigma^2$. 
Here, $m$ can be a complicated function like a neural network. 
The conditional coding length yields:
\begin{equation}
        -\log_2(\mu(y|x)) = \frac{1}{2}\log_2(2\pi \sigma^2) + \frac{(y-m(x))^2}{2\sigma^2\ln(2)}
    \label{eq:GaussianSquareLoss}
\end{equation}
When setting $\sigma^2$ and $\ell(\mu)$ constant, then the argmin of Eq.~\eqref{eq:descriptionLength} with Eq.~\eqref{eq:GaussianSquareLoss} reduces to a standard regularized MSE loss up to scaling and additive constants, revealing MSE as a special case of description length minimization. 
The full framework allows joint optimization over both $\theta$ and $\sigma$, which we prove in App.~\ref{app:refinedSquareLoss} to recover both the true function and noise level in the large-sample limit.
When $m$ is linear, this further reduces to the compressive sensing (CS) framework \cite{foucart_mathematical_2013}, where $\ell_0$ regularization provably enables sparse recovery. 
Eq.~\eqref{eq:GaussianSquareLoss} extends the CS problem to nonlinear $m$, allowing us to test whether the sample efficiency predictions from Sec.~\ref{sec:preliminaries} hold when the true data-generating function is a neural network.
To test these predictions, we introduce a \textbf{teacher-student setup} using multi-layer perceptrons (MLPs) where the student has greater capacity than the sparse teacher (details in Sec.~\ref{sec:stExperiments}).
These networks act as functions that compute the mean value $m(x)$ in the code length formula Eq.~\eqref{eq:GaussianSquareLoss} and the Solomonoff induction theorems of Sec.~\ref{sec:preliminaries} hence apply to them. 
We test these predictions empirically in Sec.~\ref{sec:stExperiments} by training student networks with varying regularization strengths across different dataset sizes and noise levels.

\subsection{Probabilistic reformulation of \texorpdfstring{$\ell_0$}{L0} regularization}
\label{sec:probabilistic-reformulations}

We review and refine probabilistic reformulations of $\ell_0$ regularized optimization, which constitute an important distinct line of methods in pruning research. We contribute to this line with a refinement that does not require Monte Carlo sampling.
In probabilistic reformulations, one rewrites $\theta = wz$, where $\theta,w\in\mathbb{R}^n$, $z\in\{0,1\}^n$ and $wz$ denotes componentwise multiplication. 
With this reparameterization, $\ell_0(\theta)$ reduces to $\ell_0(z)=\sum_i z_i$. 
Further, let $\pi_\gamma$ denote the family of all Bernoulli distributions over $\{0,1\}^n$, parameterized by $\gamma \in [0,1]^n$. 
The key step then is to rewrite objective Eq.~\eqref{eq:l0objective} as a probabilistic expectation $\mathbb{E}_{z\sim \pi_\gamma}[\alpha \ell_0(z) + \mathcal{L}(f_{wz}, x)]$. 
This reformulation was previously established for the linear case in \cite{yin2020probabilistic}.
In \cite{louizos2017learning}, the same reformulation was adopted for the nonlinear case Eq.~\eqref{eq:l0objective} but without proof. 
We provide a clean proof in App.~\ref{proof:prop:probRewrite} for the following general statement:
\begin{proposition}
    \label{prop:probRewrite}
    Let $\pi_\gamma$ be the family of all Bernoulli distributions over $\{0,1\}^n$ with parameter $\gamma\in[0,1]^n$ and let $g:\{0,1\}^n\to \mathbb{R}$ be any function. Then
    \begin{equation}
            \min_{\gamma} \mathbb{E}_{z\sim \pi_\gamma}[g(z)] = \min_z g(z)
    \end{equation}
    and at any minimum $\gamma_i$ is either $1$ or $0$ $~\forall i\in\{1,\cdots,n\}$.
\end{proposition}

The challenge in applying this powerful proposition is the resulting combinatorial complexity since writing out the expected value results in a sum with $2^n$ terms if $z\in\{0,1\}^n$. 
To evade this problem, Monte-Carlo sampling with low-variance estimators is used in \cite{yin2020probabilistic,dong2020disarm,kunes2023gradient}.
In \cite{louizos2017learning}, $z\in\{0,1\}$ is instead smoothed to a variable in $[0,1]^n$ and 
a Monte-Carlo estimator
is derived by employing a reparameterization as in \cite{kingma2013auto}.

We present a method that does not require any Monte-Carlo sampling.
We begin by noting that the expectation of a sum linear in $z$ is equal to the sum over $\gamma$ yielding
\begin{equation}
        \mathbb{E}_{z \sim \pi_\gamma}[\ell_0(z)] = \gamma_1 + \dots +  \gamma_n.
    \label{eq:l0probterm}
\end{equation}
This result was already used in \cite{louizos2017learning} to simplify the $\ell_0$ term in Eq.~\eqref{eq:l0objective}. 
By contrast, the non-linear term $\mathbb{E}_{z\sim \pi_\gamma}[\mathcal{L}(f_{wz}, x)]$ does not admit such simplification, as the loss couples all components of $z$. 

For the next step, we build on a result of \cite{barth2025probabilistic_and_nonlinear_compressive_sensing}:
\begin{equation}
    \label{lem:quadratic}
        \mathbb{E}_{z\sim\pi_\gamma} \bigg[\sum_{i}\big(y_{i} - \sum_j X_{ij}w_jz_j\big)^2\bigg] 
        = \sum_{i}\big(y_{i} - \sum_j X_{ij}w_j\gamma_j\big)^2 
        + \sum_{i,j} X_{ij}^2 w_j^2 \gamma_j (1-\gamma_j)
\end{equation}

This equation provides a way to substantially reduce the number of summands compared to the naive expectation in the case of a linear network.
However, this is still insufficient for handling the fully-nonlinear objective Eq.~\eqref{eq:l0objective}. 

To address this, we rewrite the problem as a constrained optimization problem. 
The minimum of $L_\theta(x)$ equals:
\begin{equation}
        \min_{\theta,w,z}\left\{ \alpha \ell_0(z) + \mathcal{L}(f_\theta, x) \big|\theta=wz \right\} 
        = \min_{\theta,w,z}\max_u\left\{ \alpha \ell_0(z) + \mathcal{L}(f_\theta, x) + u \cdot (\theta-wz)^2 \right\}
    \label{eq:constrained}
\end{equation}
The maximum over $u$ forces $\theta$ to equal $wz$ at the global optimum\footnote{Here $u\cdot(\theta-wz)^2$ denotes the dot product of $u\in\mathbb{R}^n$ with the componentwise square of $(\theta-wz)$; juxtaposed variables such as $wz$ denote componentwise multiplication.}, while $z$ appears at most quadratically. 
Applying Prop.~\ref{prop:probRewrite} together with Eq.~\eqref{eq:l0probterm} and Eq.~\eqref{eq:constrained} yields the following corollary:
\begin{corollary}
    \label{cor:probNonlinear}
    \begin{equation}
            \min_\theta L_\theta(x) 
            = \min_{\theta,w,\gamma}\max_u\Big\{ \alpha \sum \gamma_i + \mathcal{L}(f_\theta, x) 
            + u\cdot  (\theta-w\gamma)^2 + u\cdot(w^2\gamma(1-\gamma)) \Big\}
        \label{eq:nonlinearProb}
    \end{equation}
\end{corollary}
In App.~\ref{sec:squareTerms} we explain why a quadratic 
constraint, and thus Eq.~\eqref{lem:quadratic}, is crucial for the procedure to work.
We optimize this minimax objective with alternating gradient descent and ascent steps and list alternative strategies at the end of Section \ref{sec:ImageClassifierExperiments}. 
 
We call our method \textbf{Probabilistic Minimax Pruning (PMMP)}, reflecting 
the probabilistic reformulation in Prop.~\ref{prop:probRewrite} 
as well as the form of a minimax objective in Corollary~\ref{cor:probNonlinear}.

Finally, Eq.~\eqref{lem:quadratic} also allows to construct a \textbf{Layerwise Pruning} method:
For each layer, one can replace $X_{ij}$ in Eq.~\eqref{lem:quadratic} by input batches to that layer and promote $w_jz_j$ to matrices $w_{jk}z_{jk}$, in order to find the sparsest possible matrix that produces similar output batches $y_{ik}$. 
Details are given in App.~\ref{app:layerwisePruning} and ablation studies, that investigate its combination with our other methods in App.~\ref{app:layerwisePruningAblation}. 

\subsection{Smooth reformulations of \texorpdfstring{$\ell_0$}{L0} regularization}
\label{sec:smooth-reformulations}

\textbf{Differentiable Relaxation of $\ell_0$ Regularization (DRR).} 
A complementary approach to the probabilistic optimization methods of Eq.~\eqref{eq:l0objective} 
is to directly approximate $\ell_0(\theta)$ by some piecewise smooth function. 
The most compelling such approach we found was described in \cite{oliveira2024compression}, who use an approximation that goes back to \cite{mangasarian1996machine, bradley1998feature, gu2009l0}. 
The simple but effective idea is to write $ \ell_0(\theta) \approx \sum_i (1-e^{-\beta |\theta_i|})$.
The full optimization objective as proposed in \cite{oliveira2024compression} reads,
        $\min_\theta \mathcal{L}(f_\theta, x) + \alpha \sum_i (1-e^{-\beta |\theta_i|}) + \rho ||\theta||_2^2$.
The authors further propose to use layerwise normalization of the hyperparameters $\alpha$ and $\rho$.
The obvious advantage of this differentiable approximation of the $\ell_0$ norm is that it allows direct application of gradient descent. 
However, the method is computationally costly because it comes with three hyperparameters, $\alpha$, $\beta$, and $\rho$, which are substantially harder to fine-tune than just one. 
Additionally, parameters do not become exactly $0$ as they do in probabilistic methods, and thus one must also fix a pruning hyperparameter $\epsilon$ below which all parameters are eliminated (magnitude pruning). 
To improve the method, we investigated how to reduce the computational complexity involved in choosing these parameters. We conducted ablation studies summarized in App.~\ref{app_ablation_studies}, informing the following suggestions:
(1) contrary to \cite{oliveira2024compression}, we find that additional $\ell_2$ regularization is usually not beneficial, allowing us to drop the $\rho ||\theta||_2^2$ term in most cases; (2) we find that increasing the regularization strength $\alpha$ during training can yield lower loss and more stable outcomes. 
Additionally, we strengthen the method with several complementary techniques described in Sec.~\ref{sec:supportingMethods}, namely Random Gradient Pruning and TAMADE, which improve robustness and reduce hyperparameter sensitivity.
Since the authors did not provide a name, we call the refined method Differentiable Relaxation of $\ell_0$ Regularization (DRR) and refer to their version as DRR-O (O for ``original'').

\textbf{Relaxed $\ell_1$ regularization (R-L1).} We introduce $\ell_1$ regularization as a second smooth approximation. Its advantage is that the piecewise gradients equal the sign of the parameters and are therefore independent of their magnitudes.
Hence, all parameters are pulled toward zero with equal force.
While $\ell_1$ regularization is known to induce sparsity in compressive sensing \cite{tibshirani1996regression,foucart_mathematical_2013}, its connection to model complexity is theoretically underexplored in the nonlinear setting. 
A known disadvantage of $\ell_1$ regularization is shrinkage bias: even parameters that should remain nonzero are damped toward zero.
While this can be beneficial in high-noise settings, it generally prevents consistent parameter estimation.
This observation motivated the relaxed lasso method \cite{meinshausen2007relaxed}, whose simplified form \cite{hastie2017extended} trains with an $\ell_1$ penalty until the active set stabilizes and then removes the penalty to finetune the remaining parameters.
This strategy has also been adopted for pruning neural networks \cite{sahoo2018learning}.
We refer to this method as Relaxed $\ell_1$ regularization (R-L1).
As with DRR, we improve R-L1 with the techniques described in Sec.~\ref{sec:supportingMethods}.

\subsection{Complementary Methods}
\label{sec:supportingMethods}
\label{sec:random-gradient-pruning}
\label{sec:adaptive-mask-determination}

\textbf{Threshold Adaptive Mask Determination.} The smooth reformulations introduced in the last subsection do not set parameters exactly to zero during optimization. 
This necessitates an additional magnitude pruning step with threshold hyperparameter $\epsilon$. This threshold is coupled with the regularization strength $\alpha$ 
which complicates hyperparameter optimization.
We address this problem by introducing our first complementary method Threshold Adaptive Mask Determination (TAMADE).
It makes use of two monotone relationships: 
(1) increasing $\epsilon$ increases the number of parameters set to zero during pruning, and 
(2) beyond a certain threshold, model accuracy is non-increasing as a function of deleted parameters. 
These relationships allow us to optimize $\epsilon$ efficiently through a variant of binary search.
In practice, we determine $\epsilon$ as the largest threshold such that the 
accuracy of the model does not decrease by more than $\delta > 0$ (in absolute terms) or such that 
the loss does not increase by more than $\delta > 0$ (in relative terms). 
Pseudocode is given in App.~\ref{sec:tamade-algo}.
While the parameter $\epsilon$ needs to be optimized during the hyperparameter tuning phase and is highly coupled with other parameters, the parameter $\delta$ can be chosen according to application concerns and does not exhibit this coupling. Therefore, TAMADE effectively eliminates one dimension in the hyperparameter search problem.

\textbf{Random Gradient Pruning.} Many pruning methods leave spurious weights behind that do not contribute to the function that the network computes. For example, all outgoing connections of some neuron might have been pruned away, while incoming connections to the neuron remain. 
An example image is given in Fig.~\ref{fig:rgpIllustration} in App.~\ref{sec:rgpIllustration}. 
To solve this problem, we propose the Random Gradient Pruning method. 
It is infeasible to directly compute all elements $x$ in the domain of a parameterized function $f_\theta$ for which changes in $\theta$ leave all values $f_\theta(x)$ invariant. 
However, if the derivative of a function $\partial f_\theta(x)/\partial \theta_i$ is zero for all $x$, then $f_\theta$ does not depend on $\theta_i$. For a specific $x$, a vanishing gradient of $f_\theta(x)$ could simply indicate a local minimum with respect to $\theta_i$. Nevertheless, if we sample a \textit{random} element $z$ and assume that all important parameters in $f_\theta$ can influence its value at $z$, then we can detect dependence of $f_\theta$ on $\theta_i$ with high probability. In practice, when $f_\theta$ is a vector-valued neural network, we sample a random pair $(y,z)$ and compute
$g := \frac{\partial}{\partial \theta} ||y - f_\theta(z)||^2$.
After this, we set all $\theta_i$ to $0$ where $g_i$ is \emph{exactly} equal to $0$. 
This works well in practice to eliminate all and only the spurious weights with only a single backward pass. 


\section{Experiments}
\label{experiments}

\textbf{Main questions.} 
To achieve the aim described in Sec.~\ref{Introduction}, we answer the following empirical questions motivated by the theory developed in Sec.~\ref{theory}:
(Q1) When data is limited, does compression improve generalization and sample efficiency relative to unregularized models?
(Q2) Which compression methods reduce model size most effectively without degrading performance across a wide variety of model architectures and datasets?
(Q3) Do these methods outperform conventional compression algorithms and unregularized models?

\textbf{Computational details.} The optimal choice of hyperparameters for each method is determined through ablation studies in App.~\ref{app_ablation_studies}.  
We employ SGD with momentum for ImageNet and the ADAM optimizer \cite{kingma2014adam} for all other simulations. We divide data into training-, (test-), and validation sets. For MNIST, CIFAR, and the teacher-student experiments, we use a refined convergence criterion for early stopping (see App.~\ref{app:convergence}).
In App.~\ref{sec:computationalResources} we discuss computational resources and wallclock times, showing that
all three methods (R-L1, DRR, and PMMP) exhibit excellent computational efficiency and minimal overhead compared to unregularized training across all models and settings.

\subsection{Teacher-Student experiments}
\label{sec:stExperiments}

We empirically test the sample-efficiency predictions described in Sec.~\ref{theory} using a controlled teacher-student setting, whose loss function was related to MDL in Sec.~\ref{sec:mseandteacherstudentsetup}.
The goal is to isolate the effect of explicit description length regularization when the data-generating process itself has low description complexity.
In this setting, both teacher and student are MLPs with identical input and output dimensions but different hidden-layer widths.
The teacher is sparse in the sense that its hidden layers have fewer dimensions than those of the student. 
For our simulations the respective dimensions for teacher and student are [2, 25, 25, 1] and [2, 5, 8, 1]. 
Training data is generated by sampling inputs and drawing targets from a Gaussian distribution whose mean is given by the teacher network and the standard deviation $\sigma$ is a parameter.
Under this setup, Solomonoff induction and MDL theory predict that regularization toward lower description length should improve generalization of the student.

\begin{figure*}[ht]
    \centering
    \begin{subfigure}[t]{0.32\textwidth}
        \includegraphics[width=\textwidth]{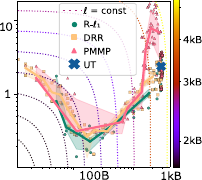}
        \caption{Teacher-Student (loss vs. size)}
        \label{fig:student-teacher-losses:b}
    \end{subfigure}
    \hfill
    \begin{subfigure}[t]{0.32\textwidth}
        \includegraphics[width=\linewidth]{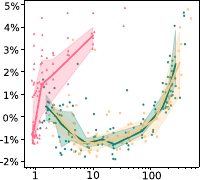}
        \caption{Cifar (EI vs. CR)}
        \label{fig:cifara}
    \end{subfigure}
    \hfill
    \begin{subfigure}[t]{0.32\textwidth}
        \includegraphics[width=\linewidth]{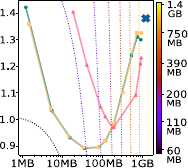}
        \caption{Transformer (loss vs. size)}
        \label{fig:transformer}
    \end{subfigure}
    \caption{{\bf Test performance vs. model size} for different machine learning problems. {\bf Left:} teacher-student setting `test loss' vs. `model size' with color-coded isolines denoting the dataset description length. {\bf Middle:} `Error increase' (EI) over baseline accuracy vs. `compression rate' (CR) \texttt{full-model-size/pruned-model-size} for the image classification problem cifar. {\bf Right:} `Test loss' vs. `model size' for a 308M parameter transformer trained on 300MB of English wikipedia text; isolines as before. Curves and error bars for Fig.~\ref{fig:student-teacher-losses:b} and \ref{fig:cifara} were computed as described in App.~\ref{app:TeacherStudentResultSummary}. For Fig.~\ref{fig:student-teacher-losses:b}, we sample data with $\sigma=0.08$ and minimize the loss \eqref{eq:GaussianSquareLoss}, see App.~\ref{app:HyperparametersForMethodComparison} for more details.}
    \label{fig:student-teacher-losses}
    \vspace{-1mm}
    \label{fig:plots}
\end{figure*}

Fig.~\ref{fig:student-teacher-losses:b} shows test loss as a function of model byte size (controlled through regularization strength). 
Colored isolines indicate total description length, computed by encoding the trained weights in sparse matrix format and encoding the dataset using the trained student.
The test loss exhibits a U-shaped curve, where the minimum marks the point where least overfitting occurs.
We compare more dataset sizes and noise-level variations in App.~\ref{app:TeacherStudentResultSummary}-Fig.~\ref{fig:student-teacher-losses:app} to study how the optimal trade-off between model complexity and predictive accuracy shifts with data availability. We also provide a quantitative summary of optimal description length results for different settings in form of Tab.~\ref{teacher-student-description-length}.
We find that for small datasets, reduced model size yields substantially lower test loss, 
providing evidence for an affirmative answer to (Q1).
For larger datasets, the U-curve is tilted, favoring bigger models as expected. 
The description length isolines interestingly show that the points of smallest loss are slightly shifted towards bigger model sizes compared to the points of optimal compression. 
Our observations 
give empirical evidence for the prediction of Solomonoff induction and its MDL approximation that regularized models exhibit more sample-efficient convergence. 

\subsection{Image classifier experiments}
\label{sec:ImageClassifierExperiments}

Image classification benchmarks provide a well-established testbed for evaluating $\ell_0$ regularization and pruning for nonlinear neural networks.
We show results on CIFAR \cite{krizhevsky2009learning} with VGG-16-512 in Tab.~\ref{tab:cifar-vgg} and on ImageNet-1k \cite{deng2009imagenet} with ResNet-50 in Tab.~\ref{tab:imagenet-resnet-50}. Additional results on MNIST \cite{lecun1998gradient} with LeNet-300-100 and LeNet-5-Caffe are given in App.~\ref{sec:classifier-benchmark-tables}. 
In all tables, error increase (EI) is the difference between the accuracy of the unregularized and the regularized model (lower is better) and CR is the ratio of non-zero parameters before and after training (higher is better).

\begin{table}[ht]
    \begin{minipage}[t]{0.48\textwidth}
        \caption{Pruning on CIFAR with VGG-16-512.
        }
        \label{tab:cifar-vgg}
        \centering
        \vspace{0.25cm}
        \begin{small}
            \begin{tabular}{lcc}
                \toprule
                \textbf{Method}    & \textbf{EI} & \textbf{CR} \\
                \midrule
                \mbox{SSS \cite{huang2018dataSSS}} & 0.00 & 1.67 \\
                \mbox{EConvNets \cite{li2016pruningEConvNet}} & -0.15 & 2.78 \\
                \mbox{VAR-CONV \cite{zhao2019variationalVARCONV}} & 0.07 & 3.75 \\
                \mbox{PP \cite{singh2019play}} & 0.14 & 6.43 \\
                \mbox{RFP \cite{AYINDE2019148RFP}} & 0.72 & 9.52 \\
                \mbox{DRR-O \cite{oliveira2024compression}} & 0.70 & 15 \\
                \mbox{SparseVD \cite{molchanov2017variationalSparseVD}} & 0.00 & 48 \\
                \mbox{CPA \cite{phan2020pruning}} & -0.01 & 56 \\
                \midrule
                \mbox{DRR (ours)} & -0.89 & \textbf{82} \\
                \mbox{R-L1 (ours)} & -0.73 & 71 \\
                \mbox{PMMP (ours)} & 3.88 & 34 \\
                \bottomrule
            \end{tabular}
        \end{small}
    \end{minipage}        
    \hfill
    \begin{minipage}[t]{0.48\textwidth}
        \caption{Pruning on ImageNet with ResNet-50.}
        \label{tab:imagenet-resnet-50}
        \centering
        \vspace{0.25cm}
        \begin{small}
            \begin{tabular}{lcc}
                \toprule
                \textbf{Method}    & \textbf{EI} & \textbf{CR} \\
                \midrule
                \mbox{PP \cite{singh2019play}} & $>$ 0.2  & $<$ 1.7 \\ 
                \mbox{SSS \cite{huang2018dataSSS}} & 0.7 (1.94) & 1.008 (1.4) \\
                \mbox{BoundedNorm \cite{mummadi2019group}} & 0.5 & 1.2 \\
                \mbox{ThiNet \cite{ThiNet}} & 0.8 & 1.5 \\
                \mbox{ChannelPruning \cite{zhuang2018discrimination}} & 1.1 & 2.1 \\
                \mbox{Autoprune \cite{xiao2019autoprune}} & 0.4 & 2.2 \\
                \mbox{VAR-CONV \cite{zhao2019variationalVARCONV}} & -0.1 & 2.5 \\
                \mbox{GSM \cite{ding2019globalGSM}} & 0.4 & 4.0 \\
                \midrule
                \mbox{DRR (ours)} & -1.2 & {\bf 8.0} \\
                \mbox{R-L1 (ours)} & 1.1 & 6.8 \\ 
                \mbox{PMMP (ours)} &  1.5& 3.5 \\
                \bottomrule
            \end{tabular}
        \end{small}
    \end{minipage}
    \vspace{-1mm}
\end{table}

Across all settings, our methods achieve very high compression rates with minimal or even negative error increase. Our smooth reformulations achieve best-in-class results. 
Among these, R-L1 offers greater simplicity, as DRR requires an additional hyperparameter $\beta$, but DRR achieves better performance.
In Fig.~\ref{fig:cifara}, we show how DRR and R-L1 do not only maintain but even increase accuracy while lowering model size.
This complements the experiments in Sec.~\ref{sec:stExperiments} (reconfirming (Q1)), consistent with our main point that compression acts as an explicit inductive bias against overfitting.

PMMP has weaker performance, possibly because minimax objectives have a very challenging optimization landscape \cite{minimaxHard2021Grimmer}. 
We explored two remedies. First, we implemented ADMM-based optimization \cite{boyd2011distributed}, see App.~\ref{app:admm} for detailed derivations of the update equations, which might be of independent interest.  
Additionally, we implemented ``fictitious play'' \cite{brown1949some,brown1951iterative} for PMMP, which is a game theoretic strategy still used by modern algorithms \cite{lee2020efficientexplorationstatemarginal}, and show results for our \textit{fictitious probabilistic pruning} (FPP) method in App.~\ref{sec:classifier-benchmark-figures}. 
Unfortunately, neither of these approaches resolved the issue. 
As our main aim is not to promote a specific method but to identify the best practical approach for achieving information theoretical goals, we refrained from exploring further alternatives in this paper.
Nevertheless, PMMP still matches other probabilistic approaches such as $L_0$-HC \cite{louizos2017learning} (Tab.~\ref{tab:mnist-lenet300}) while improving their computational efficiency 
and new minimax techniques, such as \cite{mazumdarJordan2025finding}, may improve the performance of PMMP.
DRR and R-L1 thus constitute our first answer to (Q2).

\subsection{Transformer experiments}
\label{sec:transformerExperiments}

We study whether explicit description-length regularization improves compression and generalization in neural language models, compared to both unregularized training and conventional compression algorithms.
This setting directly tests the MDL perspective introduced in Sec.~\ref{theory} in a modern sequence-prediction context.
Our experiments closely follow the setup of \cite{DeletangICLR2024}, who showed that language models act as state-of-the-art compressors.
We use Wikipedia text (subsets of Wiki-40B \cite{guo2020wiki}) and a decoder-only transformer trained by next-token prediction.
The key difference is that we explicitly optimize description length during training by regularizing the model parameters using PMMP, DRR, and R-L1, rather than evaluating compression only after unregularized training.
We focus on dataset sizes [300MB, 1.2GB, 6.2GB] for which the transformer size (308M parameters, approximately 1.2GB) is not negligible relative to the data.
The resulting dataset-model size ratios are comparable to those used in modern large language models such as LLaMA \cite{touvron2023llama} or DeepSeek-V3 \cite{deepseekai2024deepSeekV3technicalReport}. 
In fact, the largest dataset size corresponds to the 20-tokens-per-parameter rule from \cite{hoffmannTrainingComputeOptimalLarge}.

\begin{table}[ht]
    \centering
    \vspace{-2.9mm}
    \caption{{\bf Description length in MB} for different methods and dataset sizes. Conventional compressors: LZMA2 \cite{LZMA} and ZSTD \cite{Collet2021zstd}; UT\texttt{xx}: unregularized transformer with \texttt{xx} million parameters.}
    \label{descLength}
    \begin{small}
    \begin{tabular}{c|cc|ccc|ccc}
        \toprule
        \textbf{Dataset Size} & LZMA2 & ZSTD & UT308 & UT25 & UT4 & PMMP & DRR & R-L1 \\
        \midrule
        300   & 241  & 195  & 1270 & 146 & 70 & 94 & {\bf 61} & {\bf 61} \\
        1232  & 992  & 802  & 1410 & 306 & 254& 299 & 239 & {\bf 214} \\
        6160  & 4958 & 4010 & 2060 & 1080 & 1130 & 1220 & 973 & {\bf 971} \\
        \bottomrule
    \end{tabular}
    \end{small}
\end{table}

Tab.~\ref{descLength} compares conventional compressors, unregularized transformers and our regularized models (all compress chunks of 512 bytes). The transformers compress the data offline via arithmetic coding (cf.~Sec.~\ref{theory}). 
DRR and R-L1 achieve the best performance, while PMMP performs slightly worse.
Overall, Tab.~\ref{descLength} extends the results of \cite{DeletangICLR2024} to models that explicitly minimize description length during training and shows that such $\ell_0$ regularization consistently outperforms conventional compression algorithms and unregularized transformer baselines. It provides further evidence that R-L1 and DRR are good answers to (Q2) and establishes an affirmative answer to (Q3).
As mentioned in Sec.~\ref{sec:relatedWork}, an alternative online approach to data compression is to integrate the area under the entropy-loss curve as laid out in \cite{JackRaeVideo}. 
However, this does not result in better compression unless the loss converges rapidly. 
As we demonstrate for LLaMA in App.~\ref{app:llama_analysis}, this is typically not the case. On top of that, our compression methods yield significantly smaller models for inference. 

Finally, Fig.~\ref{fig:transformer} shows once more the characteristic U-curve demonstrating that our compression methods improve generalization (reconfirming (Q1)). 
Detailed loss-model-size plots and an extended qualitative analysis are provided in App.~\ref{app:TransformerStudies}. A very important effect visible in Fig.~\ref{fig:transformerTestLossVsModelSize} is that smaller unregularized models underperform regularized models, even when both yield the same final parameter count.
This suggests that fixing a reduced architecture \emph{a priori} restricts model training to a suboptimal parameter subspace, whereas compression-regularization improves generalization by optimizing in a much larger and more expressive space, while still resulting in compact models.


\section{Conclusion and Limitations}

We studied neural network compression through the lens of description length minimization and $\ell_0$ regularization, framing compression as an explicit inductive bias. 
Across architectures and datasets, compression-regularized training substantially reduces model size while preserving or, in some regimes, improving generalization performance, consistent with predictions from Solomonoff induction and MDL.
Among the optimization strategies considered, smooth relaxations of the $\ell_0$ objective provide the most favorable empirical trade-off between compression and optimization stability.
Presumably due to the involved minimax optimization landscape, PMMP exhibits lower performance. However, it offers an efficient alternative to sampling-based methods and clarifies the structure of $\ell_0$ regularized objectives, thereby developing the line of probabilistic pruning approaches.

\textbf{Limitations.} We approximate description length using norm-based penalties such as $\ell_0$ and $\ell_1$, which provide a coarse but computationally tractable measure of model complexity.
This approximation does not capture higher-order regularities such as structured sparsity, weight symmetries, or algorithmic patterns in parameters, and optimizing $\ell_0$ regularized objectives in large nonlinear models remains challenging.
Nevertheless, as discussed in Sec.~\ref{sec:tractableApproximation}, parameter-count-based penalties can be bootstrapped into a hierarchy of priors that encode increasingly rich distributional information.
A promising direction for future work is to extend this hierarchy toward more general representations of data-generating processes, while remaining compatible with gradient-based optimization.

An \textbf{impact statement} is provided in Appendix \ref{app:societalImpact}.

\newpage
\begin{ack}
We thank J\"urgen Jost, Rostislav Matveev, Guido Mont\'ufar and Marcus Hutter for stimulating discussions on the topic of this article.
We gratefully acknowledge financial support from the Max Planck Institute for Mathematics in the Sciences (MPI MiS) and the German Academic Scholarship Foundation, as well as computational resources provided by MPI MiS, the Max Planck Computing and Data Facility and the Max Planck Institute of Molecular Cell Biology and Genetics / Center for Systems Biology Dresden.
The authors declare that they have no competing interests (related financial activities outside the submitted work).
\end{ack}

\vspace{0.5cm}

\small
\bibliography{bib}

@misc{jiang2024mixtralexperts,
      title={Mixtral of Experts}, 
      author={Albert Q. Jiang and Alexandre Sablayrolles and Antoine Roux and Arthur Mensch and Blanche Savary and Chris Bamford and Devendra Singh Chaplot and Diego de las Casas and Emma Bou Hanna and Florian Bressand and Gianna Lengyel and Guillaume Bour and Guillaume Lample and Lélio Renard Lavaud and Lucile Saulnier and Marie-Anne Lachaux and Pierre Stock and Sandeep Subramanian and Sophia Yang and Szymon Antoniak and Teven Le Scao and Théophile Gervet and Thibaut Lavril and Thomas Wang and Timothée Lacroix and William El Sayed},
      year={2024},
      eprint={2401.04088},
      archivePrefix={arXiv},
      primaryClass={cs.LG},
      url={https://arxiv.org/abs/2401.04088}, 
}

@misc{minimaxHard2021Grimmer,
      title={Limiting Behaviors of Nonconvex-Nonconcave Minimax Optimization via Continuous-Time Systems}, 
      author={Benjamin Grimmer and Haihao Lu and Pratik Worah and Vahab Mirrokni},
      year={2021},
      eprint={2010.10628},
      archivePrefix={arXiv},
      primaryClass={math.OC},
      url={https://arxiv.org/abs/2010.10628}, 
}

@article{mazumdarJordan2025finding,
  title={On finding local nash equilibria (and only local nash equilibria) in zero-sum games},
  author={Mazumdar, Eric and Sastry, S Shankar and Jordan, Michael I},
  journal={ACM/IMS Journal of Data Science},
  volume={2},
  number={2},
  pages={1--26},
  year={2025},
  publisher={ACM New York, NY}
}

@misc{lee2020efficientexplorationstatemarginal,
      title={Efficient Exploration via State Marginal Matching}, 
      author={Lisa Lee and Benjamin Eysenbach and Emilio Parisotto and Eric Xing and Sergey Levine and Ruslan Salakhutdinov},
      year={2020},
      eprint={1906.05274},
      archivePrefix={arXiv},
      primaryClass={cs.LG},
      url={https://arxiv.org/abs/1906.05274}, 
}

@article{deletang2023language,
  title={Language modeling is compression},
  author={Del{\'e}tang, Gr{\'e}goire and Ruoss, Anian and Duquenne, Paul-Ambroise and Catt, Elliot and Genewein, Tim and Mattern, Christopher and Grau-Moya, Jordi and Wenliang, Li Kevin and Aitchison, Matthew and Orseau, Laurent and others},
  journal={arXiv preprint arXiv:2309.10668},
  year={2023}
}

@inproceedings{DeletangICLR2024,
 author = {Deletang, Gregoire and Ruoss, Anian and Duquenne, Paul-Ambroise and Catt, Elliot and Genewein, Tim and Mattern, Christopher and Grau-Moya, Jordi and Wenliang, Li Kevin and Aitchison, Matthew and Orseau, Laurent and Hutter, Marcus and Veness, Joel},
 booktitle = {International Conference on Learning Representations},
 pages = {14165--14181},
 title = {Language Modeling Is Compression},
 volume = {2024},
 year = {2024}
}

@inproceedings{vaswani2017attention,
  title = {Attention Is All You Need},
  booktitle = {Advances in Neural Information Processing Systems},
  author = {Vaswani, Ashish and Shazeer, Noam and Parmar, Niki and Uszkoreit, Jakob and Jones, Llion and Gomez, Aidan N and Kaiser, {\L}ukasz and Polosukhin, Illia},
  editor = {Guyon, I. and Luxburg, U. Von and Bengio, S. and Wallach, H. and Fergus, R. and Vishwanathan, S. and Garnett, R.},
  year = 2017,
  volume = {30},
  publisher = {Curran Associates, Inc.}
}

@article{louizos2017learning,
  title={Learning sparse neural networks through {L}0 regularization},
  author={Louizos, Christos and Welling, Max and Kingma, Diederik P},
  journal={arXiv preprint arXiv:1712.01312},
  year={2017}
}

@article{yin2020probabilistic,
  title={Probabilistic best subset selection via gradient-based optimization},
  author={Yin, Mingzhang and Ho, Nhat and Yan, Bowei and Qian, Xiaoning and Zhou, Mingyuan},
  journal={arXiv preprint arXiv:2006.06448},
  year={2020}
}

@inproceedings{kunes2023gradient,
  title={Gradient estimation for binary latent variables via gradient variance clipping},
  author={Kunes, Russell Z and Yin, Mingzhang and Land, Max and Haviv, Doron and Pe'er, Dana and Tavar{\'e}, Simon},
  booktitle={Proceedings of the AAAI Conference on Artificial Intelligence},
  volume={37-7},
  pages={8405--8412},
  year={2023}
}

@article{lecun1989optimal,
  title={Optimal brain damage},
  author={LeCun, Yann and Denker, John and Solla, Sara},
  journal={Advances in neural information processing systems},
  volume={2},
  year={1989}
}

@book{foucart_mathematical_2013,
	address = {New York, NY},
	series = {Applied and {Numerical} {Harmonic} {Analysis}},
	title = {A {Mathematical} {Introduction} to {Compressive} {Sensing}},
	isbn = {978-0-8176-4947-0 978-0-8176-4948-7},
	url = {https://link.springer.com/10.1007/978-0-8176-4948-7},
	language = {en},
	urldate = {2023-11-06},
	publisher = {Springer New York},
	author = {Foucart, Simon and Rauhut, Holger},
	year = {2013},
	doi = {10.1007/978-0-8176-4948-7},
}

@article{dong2020disarm,
  title={DisARM: An antithetic gradient estimator for binary latent variables},
  author={Dong, Zhe and Mnih, Andriy and Tucker, George},
  journal={Advances in neural information processing systems},
  volume={33},
  pages={18637--18647},
  year={2020}
}

@inproceedings{guo2020wiki,
  title={Wiki-40b: Multilingual language model dataset},
  author={Guo, Mandy and Dai, Zihang and Vrande{\v{c}}i{\'c}, Denny and Al-Rfou, Rami},
  booktitle={Proceedings of the Twelfth Language Resources and Evaluation Conference},
  pages={2440--2452},
  year={2020}
}

@ARTICLE{RissanenArithmeticCoding,
  author={Rissanen, J. J.},
  journal={IBM Journal of Research and Development}, 
  title={Generalized Kraft Inequality and Arithmetic Coding}, 
  year={1976},
  volume={20},
  number={3},
  pages={198-203},
  doi={10.1147/rd.203.0198}
}

@techreport{brown1949some,
  title={Some notes on computation of games solutions},
  author={Brown, George W},
  year={1949}
}

@article{brown1951iterative,
  title={Iterative solution of games by fictitious play},
  author={Brown, George W},
  journal={Act. Anal. Prod Allocation},
  volume={13},
  number={1},
  pages={374},
  year={1951}
}

@misc{loshchilov2019decoupledweightdecayregularization,
      title={Decoupled Weight Decay Regularization}, 
      author={Ilya Loshchilov and Frank Hutter},
      year={2019},
      eprint={1711.05101},
      archivePrefix={arXiv},
      primaryClass={cs.LG},
      url={https://arxiv.org/abs/1711.05101}, 
}

@misc{barth2025probabilistic_and_nonlinear_compressive_sensing,
      title={Probabilistic and nonlinear compressive sensing}, 
      author={Lukas Silvester Barth and Paulo {von Petersenn}},
      year={2025},
      eprint={2509.15060},
      archivePrefix={arXiv},
      primaryClass={cs.LG},
      url={https://arxiv.org/abs/2509.15060}, 
}

@phdthesis{pasco1976source,
  title={Source coding algorithms for fast data compression},
  author={Pasco, Richard Clark},
  year={1976},
  school={Stanford University CA}
}

@misc{draxler2019essentiallybarriersneuralnetwork,
      title={Essentially No Barriers in Neural Network Energy Landscape}, 
      author={Felix Draxler and Kambis Veschgini and Manfred Salmhofer and Fred A. Hamprecht},
      year={2019},
      eprint={1803.00885},
      archivePrefix={arXiv},
      primaryClass={stat.ML},
      url={https://arxiv.org/abs/1803.00885}, 
}

@article{hastie2017extended,
  title={Extended comparisons of best subset selection, forward stepwise selection, and the lasso},
  author={Hastie, Trevor and Tibshirani, Robert and Tibshirani, Ryan J},
  journal={arXiv preprint arXiv:1707.08692},
  year={2017}
}

@article{meinshausen2007relaxed,
  title={Relaxed lasso},
  author={Meinshausen, Nicolai},
  journal={Computational Statistics \& Data Analysis},
  volume={52},
  number={1},
  pages={374--393},
  year={2007},
  publisher={Elsevier}
}

@article{kingma2014adam,
  title={Adam: A method for stochastic optimization},
  author={Kingma, Diederik P and Ba, Jimmy},
  journal={arXiv preprint arXiv:1412.6980},
  year={2014}
}

@article{shannon1948mathematical,
  title={A mathematical theory of communication},
  author={Shannon, Claude Elwood},
  journal={The Bell system technical journal},
  volume={27},
  number={3},
  pages={379--423},
  year={1948},
  publisher={Nokia Bell Labs}
}

@article{kolmogorov1963tables,
  title={On tables of random numbers},
  author={Kolmogorov, Andrei N},
  journal={Sankhy{\=a}: The Indian Journal of Statistics, Series A},
  pages={369--376},
  year={1963},
  publisher={JSTOR}
}

@book{shen2017kolmogorov,
  title={Kolmogorov complexity and algorithmic randomness},
  author={Shen, Alexander and Uspensky, Vladimir A and Vereshchagin, Nikolay},
  volume={220},
  year={2017},
  publisher={American Mathematical Soc.}
}

@article{solomonoff1964formal,
  title={A formal theory of inductive inference. Part I},
  author={Solomonoff, Ray J},
  journal={Information and control},
  volume={7},
  number={1},
  pages={1--22},
  year={1964},
  publisher={Elsevier}
}

@article{solomonoff1978complexity,
  title={Complexity-based induction systems: comparisons and convergence theorems},
  author={Solomonoff, Ray},
  journal={IEEE transactions on Information Theory},
  volume={24},
  number={4},
  pages={422--432},
  year={1978},
  publisher={IEEE}
}

@article{rissanen1978modeling,
  title={Modeling by shortest data description},
  author={Rissanen, Jorma},
  journal={Automatica},
  volume={14},
  number={5},
  pages={465--471},
  year={1978},
  publisher={Elsevier}
}

@book{grunwald2007minimum,
  title={The minimum description length principle},
  author={Gr{\"u}nwald, Peter D},
  year={2007},
  publisher={MIT press}
}

@book{hutter2005universal,
  title={Universal artificial intelligence: Sequential decisions based on algorithmic probability},
  author={Hutter, Marcus},
  year={2005},
  publisher={Springer Science \& Business Media}
}

@article{meulemans2025embedded,
  title={Embedded Universal Predictive Intelligence: a coherent framework for multi-agent learning},
  author={Meulemans, Alexander and Nasser, Rajai and Wo{\l}czyk, Maciej and Weis, Marissa A and Kobayashi, Seijin and Richards, Blake and Lajoie, Guillaume and Steger, Angelika and Hutter, Marcus and Manyika, James and others},
  journal={arXiv preprint arXiv:2511.22226},
  year={2025}
}

@article{barron2002minimum,
  title={Minimum complexity density estimation},
  author={Barron, Andrew R and Cover, Thomas M},
  journal={IEEE transactions on information theory},
  volume={37},
  number={4},
  pages={1034--1054},
  year={2002},
  publisher={IEEE}
}

@inproceedings{poland2004convergence,
  title={Convergence of discrete MDL for sequential prediction},
  author={Poland, Jan and Hutter, Marcus},
  booktitle={Learning Theory: 17th Annual Conference on Learning Theory, COLT 2004, Banff, Canada, July 1-4, 2004. Proceedings 17},
  pages={300--314},
  year={2004},
  organization={Springer}
}

@inproceedings{bornschein2023sequential,
title={Sequential Learning of Neural Networks for Prequential {MDL}},
author={Jorg Bornschein and Yazhe Li and Marcus Hutter},
booktitle={The Eleventh International Conference on Learning Representations },
year={2023},
url={https://openreview.net/forum?id=dMMPUvNSYJr}
}

@article{philip1999prequential,
  title={Prequential probability: principles and properties},
  author={Philip Dawid, A and Vovk, Vladimir G},
  journal={Bernoulli},
  volume={5},
  number={1},
  pages={125--162},
  year={1999}
}

@article{LZMA,
  title = {7z Format},
  year={2019},
  url={http://www.7-zip.org/7z.html},
  author={Igor Pavlov}
}

@article{tibshirani1996regression,
  title={Regression shrinkage and selection via the lasso},
  author={Tibshirani, Robert},
  journal={Journal of the Royal Statistical Society Series B: Statistical Methodology},
  volume={58},
  number={1},
  pages={267--288},
  year={1996},
  publisher={Oxford University Press}
}

@INPROCEEDINGS{OrthogonalMatchinPursuit1993,
  author={Pati, Y.C. and Rezaiifar, R. and Krishnaprasad, P.S.},
  booktitle={Proceedings of 27th Asilomar Conference on Signals, Systems and Computers}, 
  title={Orthogonal matching pursuit: recursive function approximation with applications to wavelet decomposition}, 
  year={1993},
  volume={},
  number={},
  pages={40-44 vol.1},
  doi={10.1109/ACSSC.1993.342465}
}

@misc{JackRaeVideo,
	author = {Jack Rae},
  year = {2023},
	title = {{Compression for AGI  | Stanford MLSys \#76 | Talk on Youtube}},
	url = {https://www.youtube.com/watch?v=dO4TPJkeaaU},
}

@inproceedings{sahoo2018learning,
  title={Learning equations for extrapolation and control},
  author={Sahoo, Subham and Lampert, Christoph and Martius, Georg},
  booktitle={International Conference on machine learning},
  pages={4442--4450},
  year={2018},
  organization={Pmlr}
}

@article{kingma2013auto,
  title={Auto-encoding variational bayes},
  author={Kingma, Diederik P},
  journal={arXiv preprint arXiv:1312.6114},
  year={2013}
}

@article{blumensath2008iterative,
  title={Iterative thresholding for sparse approximations},
  author={Blumensath, Thomas and Davies, Mike E},
  journal={Journal of Fourier analysis and Applications},
  volume={14},
  pages={629--654},
  year={2008},
  publisher={Springer}
}

@article{boyd2011distributed,
  title={Distributed optimization and statistical learning via the alternating direction method of multipliers},
  author={Boyd, Stephen and Parikh, Neal and Chu, Eric and Peleato, Borja and Eckstein, Jonathan and others},
  journal={Foundations and Trends{\textregistered} in Machine learning},
  volume={3},
  number={1},
  pages={1--122},
  year={2011},
  publisher={Now Publishers, Inc.}
}

@article{oliveira2024compression,
  title={On the compression of neural networks using l0-norm regularization and weight pruning},
  author={de Resende Oliveira, Felipe Dennis and Batista, Eduardo Luiz Ortiz and Seara, Rui},
  journal={Neural Networks},
  volume={171},
  pages={343--352},
  year={2024},
  publisher={Elsevier}
}

@article{gu2009l0,
  title={The l0 norm constraint LMS algorithm for sparse system identification},
  author={Gu, Yuantao and Jin, Jian and Mei, Shunliang},
  journal={IEEE Signal Processing Letters},
  volume={16},
  number={9},
  pages={774--777},
  year={2009},
  publisher={IEEE}
}

@inproceedings{bradley1998feature,
  title={Feature selection via concave minimization and support vector machines.},
  author={Bradley, Paul S and Mangasarian, Olvi L},
  booktitle={ICML},
  volume={98},
  pages={82--90},
  year={1998}
}

@incollection{mangasarian1996machine,
  title={Machine learning via polyhedral concave minimization},
  author={Mangasarian, Olvi L},
  booktitle={Applied Mathematics and Parallel Computing: Festschrift for Klaus Ritter},
  pages={175--188},
  year={1996},
  publisher={Springer}
}

@article{touvron2023llama,
  title={Llama: Open and efficient foundation language models},
  author={Touvron, Hugo and Lavril, Thibaut and Izacard, Gautier and Martinet, Xavier and Lachaux, Marie-Anne and Lacroix, Timoth{\'e}e and Rozi{\`e}re, Baptiste and Goyal, Naman and Hambro, Eric and Azhar, Faisal and others},
  journal={arXiv preprint arXiv:2302.13971},
  year={2023}
}

@misc{deepseekai2024deepSeekV3technicalReport,
      title={DeepSeek-V3 Technical Report}, 
      author={DeepSeek-AI},
      year={2024},
      eprint={2412.19437},
      archivePrefix={arXiv},
      primaryClass={cs.CL},
      url={https://arxiv.org/abs/2412.19437}, 
}

@inproceedings{esser2024scaling,
  title={Scaling rectified flow transformers for high-resolution image synthesis},
  author={Esser, Patrick and Kulal, Sumith and Blattmann, Andreas and Entezari, Rahim and M{\"u}ller, Jonas and Saini, Harry and Levi, Yam and Lorenz, Dominik and Sauer, Axel and Boesel, Frederic and others},
  booktitle={Forty-first International Conference on machine learning},
  year={2024}
}

@misc{imagenteamgoogle2024imagen3,
      title={Imagen 3}, 
      author={Imagen-Team-Google and : and Jason Baldridge and Jakob Bauer and Mukul Bhutani and Nicole Brichtova and Andrew Bunner and Lluis Castrejon and Kelvin Chan and Yichang Chen and Sander Dieleman and Yuqing Du and Zach Eaton-Rosen and Hongliang Fei and Nando de Freitas and Yilin Gao and Evgeny Gladchenko and Sergio Gómez Colmenarejo and Mandy Guo and Alex Haig and Will Hawkins and Hexiang Hu and Huilian Huang and Tobenna Peter Igwe and Christos Kaplanis and Siavash Khodadadeh and Yelin Kim and Ksenia Konyushkova and Karol Langner and Eric Lau and Rory Lawton and Shixin Luo and Soňa Mokrá and Henna Nandwani and Yasumasa Onoe and Aäron van den Oord and Zarana Parekh and Jordi Pont-Tuset and Hang Qi and Rui Qian and Deepak Ramachandran and Poorva Rane and Abdullah Rashwan and Ali Razavi and Robert Riachi and Hansa Srinivasan and Srivatsan Srinivasan and Robin Strudel and Benigno Uria and Oliver Wang and Su Wang and Austin Waters and Chris Wolff and Auriel Wright and Zhisheng Xiao and Hao Xiong and Keyang Xu and Marc van Zee and Junlin Zhang and Katie Zhang and Wenlei Zhou and Konrad Zolna and Ola Aboubakar and Canfer Akbulut and Oscar Akerlund and Isabela Albuquerque and Nina Anderson and Marco Andreetto and Lora Aroyo and Ben Bariach and David Barker and Sherry Ben and Dana Berman and Courtney Biles and Irina Blok and Pankil Botadra and Jenny Brennan and Karla Brown and John Buckley and Rudy Bunel and Elie Bursztein and Christina Butterfield and Ben Caine and Viral Carpenter and Norman Casagrande and Ming-Wei Chang and Solomon Chang and Shamik Chaudhuri and Tony Chen and John Choi and Dmitry Churbanau and Nathan Clement and Matan Cohen and Forrester Cole and Mikhail Dektiarev and Vincent Du and Praneet Dutta and Tom Eccles and Ndidi Elue and Ashley Feden and Shlomi Fruchter and Frankie Garcia and Roopal Garg and Weina Ge and Ahmed Ghazy and Bryant Gipson and Andrew Goodman and Dawid Górny and Sven Gowal and Khyatti Gupta and Yoni Halpern and Yena Han and Susan Hao and Jamie Hayes and Jonathan Heek and Amir Hertz and Ed Hirst and Emiel Hoogeboom and Tingbo Hou and Heidi Howard and Mohamed Ibrahim and Dirichi Ike-Njoku and Joana Iljazi and Vlad Ionescu and William Isaac and Reena Jana and Gemma Jennings and Donovon Jenson and Xuhui Jia and Kerry Jones and Xiaoen Ju and Ivana Kajic and Christos Kaplanis and Burcu Karagol Ayan and Jacob Kelly and Suraj Kothawade and Christina Kouridi and Ira Ktena and Jolanda Kumakaw and Dana Kurniawan and Dmitry Lagun and Lily Lavitas and Jason Lee and Tao Li and Marco Liang and Maggie Li-Calis and Yuchi Liu and Javier Lopez Alberca and Matthieu Kim Lorrain and Peggy Lu and Kristian Lum and Yukun Ma and Chase Malik and John Mellor and Thomas Mensink and Inbar Mosseri and Tom Murray and Aida Nematzadeh and Paul Nicholas and Signe Nørly and João Gabriel Oliveira and Guillermo Ortiz-Jimenez and Michela Paganini and Tom Le Paine and Roni Paiss and Alicia Parrish and Anne Peckham and Vikas Peswani and Igor Petrovski and Tobias Pfaff and Alex Pirozhenko and Ryan Poplin and Utsav Prabhu and Yuan Qi and Matthew Rahtz and Cyrus Rashtchian and Charvi Rastogi and Amit Raul and Ali Razavi and Sylvestre-Alvise Rebuffi and Susanna Ricco and Felix Riedel and Dirk Robinson and Pankaj Rohatgi and Bill Rosgen and Sarah Rumbley and Moonkyung Ryu and Anthony Salgado and Tim Salimans and Sahil Singla and Florian Schroff and Candice Schumann and Tanmay Shah and Eleni Shaw and Gregory Shaw and Brendan Shillingford and Kaushik Shivakumar and Dennis Shtatnov and Zach Singer and Evgeny Sluzhaev and Valerii Sokolov and Thibault Sottiaux and Florian Stimberg and Brad Stone and David Stutz and Yu-Chuan Su and Eric Tabellion and Shuai Tang and David Tao and Kurt Thomas and Gregory Thornton and Andeep Toor and Cristian Udrescu and Aayush Upadhyay and Cristina Vasconcelos and Alex Vasiloff and Andrey Voynov and Amanda Walker and Luyu Wang and Miaosen Wang and Simon Wang and Stanley Wang and Qifei Wang and Yuxiao Wang and Ágoston Weisz and Olivia Wiles and Chenxia Wu and Xingyu Federico Xu and Andrew Xue and Jianbo Yang and Luo Yu and Mete Yurtoglu and Ali Zand and Han Zhang and Jiageng Zhang and Catherine Zhao and Adilet Zhaxybay and Miao Zhou and Shengqi Zhu and Zhenkai Zhu and Dawn Bloxwich and Mahyar Bordbar and Luis C. Cobo and Eli Collins and Shengyang Dai and Tulsee Doshi and Anca Dragan and Douglas Eck and Demis Hassabis and Sissie Hsiao and Tom Hume and Koray Kavukcuoglu and Helen King and Jack Krawczyk and Yeqing Li and Kathy Meier-Hellstern and Andras Orban and Yury Pinsky and Amar Subramanya and Oriol Vinyals and Ting Yu and Yori Zwols},
      year={2024},
      eprint={2408.07009},
      archivePrefix={arXiv},
      primaryClass={cs.CV},
      url={https://arxiv.org/abs/2408.07009}, 
}

@article{rae2021scaling,
  title={Scaling language models: Methods, analysis \& insights from training gopher},
  author={Rae, Jack W and Borgeaud, Sebastian and Cai, Trevor and Millican, Katie and Hoffmann, Jordan and Song, Francis and Aslanides, John and Henderson, Sarah and Ring, Roman and Young, Susannah and others},
  journal={arXiv preprint arXiv:2112.11446},
  year={2021}
}

@article{rackauckas2020universal,
  title={Universal differential equations for scientific machine learning},
  author={Rackauckas, Christopher and Ma, Yingbo and Martensen, Julius and Warner, Collin and Zubov, Kirill and Supekar, Rohit and Skinner, Dominic and Ramadhan, Ali},
  journal={arXiv preprint arXiv:2001.04385},
  year={2020}
}

@MISC{pal2023lux,
  author    = {Pal, Avik},
  title     = {Lux: Explicit Parameterization of Deep Neural Networks in Julia},
  month     = {April},
  year      = 2023,
  note      = {If you use this software, please cite it as below.},
  publisher = {Zenodo},
  version   = {v0.5.0},
  doi       = {10.5281/zenodo.7808904},
  url       = {https://doi.org/10.5281/zenodo.7808904}
}

@MISC{pal2023efficient,
  title     = {On Efficient Training \& Inference of Neural Differential Equations},
  author    = {Pal, Avik},
  year      = {2023},
  school    = {Massachusetts Institute of Technology}
}

@article{lecun1998gradient,
  title={Gradient-based learning applied to document recognition},
  author={LeCun, Yann and Bottou, L{\'e}on and Bengio, Yoshua and Haffner, Patrick},
  journal={Proceedings of the IEEE},
  volume={86},
  number={11},
  pages={2278--2324},
  year={1998},
  publisher={Ieee}
}

@article{krizhevsky2009learning,
  title={Learning multiple layers of features from tiny images},
  author={Krizhevsky, Alex and Hinton, Geoffrey and others},
  year={2009},
  journal={N.A.},
  publisher={Toronto, ON, Canada}
}

@inproceedings{phan2020pruning,
  title={Pruning Deep Neural Networks with L\_0-Constrained Optimization},
  author={Phan, Dzung T and Nguyen, Lam M and Nguyen, Nam H and Kalagnanam, Jayant R},
  booktitle={2020 IEEE International Conference on Data Mining (ICDM)},
  pages={1214--1219},
  year={2020},
  organization={IEEE}
}

@article{ullrich2017soft,
  title={Soft weight-sharing for neural network compression},
  author={Ullrich, Karen and Meeds, Edward and Welling, Max},
  journal={arXiv preprint arXiv:1702.04008},
  year={2017}
}

@article{lee2018snip,
  title={Snip: Single-shot network pruning based on connection sensitivity},
  author={Lee, Namhoon and Ajanthan, Thalaiyasingam and Torr, Philip HS},
  journal={arXiv preprint arXiv:1810.02340},
  year={2018}
}

@article{guo2016DNS,
  title={Dynamic network surgery for efficient dnns},
  author={Guo, Yiwen and Yao, Anbang and Chen, Yurong},
  journal={Advances in neural information processing systems},
  volume={29},
  year={2016}
}

@inproceedings{molchanov2017variationalSparseVD,
  title={Variational dropout sparsifies deep neural networks},
  author={Molchanov, Dmitry and Ashukha, Arsenii and Vetrov, Dmitry},
  booktitle={International conference on machine learning},
  pages={2498--2507},
  year={2017},
  organization={PMLR}
}

@article{ding2019globalGSM,
  title={Global sparse momentum sgd for pruning very deep neural networks},
  author={Ding, Xiaohan and Zhou, Xiangxin and Guo, Yuchen and Han, Jungong and Liu, Ji and others},
  journal={Advances in Neural Information Processing Systems},
  volume={32},
  year={2019}
}

@article{xiao2019autoprune,
  title={Autoprune: Automatic network pruning by regularizing auxiliary parameters},
  author={Xiao, Xia and Wang, Zigeng and Rajasekaran, Sanguthevar},
  journal={Advances in neural information processing systems},
  volume={32},
  year={2019}
}

@inproceedings{li2019L0arm,
  title={L0-ARM: Network Sparsification via Stochastic Binary Optimization},
  author={Li, Yang and Ji, Shihao},
  booktitle={Joint European Conference on machine learning and Knowledge Discovery in Databases},
  pages={432--448},
  year={2019},
  organization={Springer}
}

@inproceedings{zhang2018L0ADMM,
  title={A systematic dnn weight pruning framework using alternating direction method of multipliers},
  author={Zhang, Tianyun and Ye, Shaokai and Zhang, Kaiqi and Tang, Jian and Wen, Wujie and Fardad, Makan and Wang, Yanzhi},
  booktitle={Proceedings of the European conference on computer vision (ECCV)},
  pages={184--199},
  year={2018}
}

@inproceedings{idelbayev2022exploringL0L2LC,
  title={Exploring the Effect of L0 / L2 Regularization in Neural Network Pruning using the LC Toolkit},
  author={Idelbayev, Yerlan and Carreira-Perpi{\~n}{\'a}n, Miguel {\'A}},
  booktitle={ICASSP 2022-2022 IEEE International Conference on Acoustics, Speech and Signal Processing (ICASSP)},
  pages={3373--3377},
  year={2022},
  organization={IEEE}
}

@inproceedings{huang2018dataSSS,
  title={Data-driven sparse structure selection for deep neural networks},
  author={Huang, Zehao and Wang, Naiyan},
  booktitle={Proceedings of the European conference on computer vision (ECCV)},
  pages={304--320},
  year={2018}
}

@article{li2016pruningEConvNet,
  title={Pruning filters for efficient convnets},
  author={Li, Hao and Kadav, Asim and Durdanovic, Igor and Samet, Hanan and Graf, Hans Peter},
  journal={arXiv preprint arXiv:1608.08710},
  year={2016}
}

@inproceedings{zhao2019variationalVARCONV,
  title={Variational convolutional neural network pruning},
  author={Zhao, Chenglong and Ni, Bingbing and Zhang, Jian and Zhao, Qiwei and Zhang, Wenjun and Tian, Qi},
  booktitle={Proceedings of the IEEE/CVF conference on computer vision and pattern recognition},
  pages={2780--2789},
  year={2019}
}

@article{singh2019play,
  title={Play and prune: Adaptive filter pruning for deep model compression},
  author={Singh, Pravendra and Verma, Vinay Kumar and Rai, Piyush and Namboodiri, Vinay P},
  journal={arXiv preprint arXiv:1905.04446},
  year={2019}
}

@article{AYINDE2019148RFP,
title = {Redundant feature pruning for accelerated inference in deep neural networks},
journal = {Neural Networks},
volume = {118},
pages = {148-158},
year = {2019},
issn = {0893-6080},
doi = {https://doi.org/10.1016/j.neunet.2019.04.021},
url = {https://www.sciencedirect.com/science/article/pii/S0893608019301273},
author = {Babajide O. Ayinde and Tamer Inanc and Jacek M. Zurada}
}

@article{Edelsbrunner1983ontheshape,
  author={Edelsbrunner, H. and Kirkpatrick, D. and Seidel, R.},
  journal={IEEE Transactions on Information Theory}, 
  title={On the shape of a set of points in the plane}, 
  year={1983},
  volume={29},
  number={4},
  pages={551-559},
  doi={10.1109/TIT.1983.1056714}
}

@misc{GMT.jl,
  author       = {Joaquim F. and contributors},
  title        = {GMT.jl: Julia wrapper for the Generic Mapping Tools},
  year         = {2014--2025},
  howpublished = {online},
  url          = {https://github.com/GenericMappingTools/GMT.jl},
  note         = {Accessed: 2025-05-02}
}

@article{Collet2021zstd,
  author    = {Yann Collet and Murray Kucherawy},
  title     = {{Zstandard Compression and the 'application/zstd' Media Type}},
  howpublished = {RFC 8878 (Informational)},
  type      = {Request for Comments},
  number    = {8878},
  institution = {IETF},
  month     = {February},
  year      = {2021},
  url       = {https://www.rfc-editor.org/rfc/rfc8878.html},
  note      = {Obsoletes RFC 8478}
}

@inproceedings{mummadi2019group,
  title={Group Pruning Using a Bounded-Lp Norm for Group Gating and Regularization},
  author={Mummadi, Chaithanya Kumar and Genewein, Tim and Zhang, Dan and Brox, Thomas and Fischer, Volker},
  booktitle={German Conference on Pattern Recognition},
  pages={139--155},
  year={2019},
  organization={Springer}
}

@INPROCEEDINGS{ThiNet,
  author={Luo, Jian-Hao and Wu, Jianxin and Lin, Weiyao},
  booktitle={2017 IEEE International Conference on Computer Vision (ICCV)}, 
  title={ThiNet: A Filter Level Pruning Method for Deep Neural Network Compression}, 
  year={2017},
  volume={},
  number={},
  pages={5068-5076},
  doi={10.1109/ICCV.2017.541}}

@article{zhuang2018discrimination,
  title={Discrimination-aware channel pruning for deep neural networks},
  author={Zhuang, Zhuangwei and Tan, Mingkui and Zhuang, Bohan and Liu, Jing and Guo, Yong and Wu, Qingyao and Huang, Junzhou and Zhu, Jinhui},
  journal={Advances in neural information processing systems},
  volume={31},
  year={2018}
}

@misc{hoffmannTrainingComputeOptimalLarge,
  title = {Training {{Compute-Optimal Large Language Models}}},
  author = {Hoffmann, Jordan and Borgeaud, Sebastian and Mensch, Arthur and Buchatskaya, Elena and Cai, Trevor and Rutherford, Eliza and Casas, Diego de Las and Hendricks, Lisa Anne and Welbl, Johannes and Clark, Aidan and Hennigan, Tom and Noland, Eric and Millican, Katie},
  year = 2022,
  doi = {10.48550/arXiv.2203.15556},
}

@inproceedings{deng2009imagenet,
  title={Imagenet: A large-scale hierarchical image database},
  author={Deng, Jia and Dong, Wei and Socher, Richard and Li, Li-Jia and Li, Kai and Fei-Fei, Li},
  booktitle={2009 IEEE conference on computer vision and pattern recognition},
  pages={248--255},
  year={2009},
  organization={Ieee}
}
\bibliographystyle{plain}
\normalsize



\newpage
\onecolumn

\appendix
\section*{\LARGE{Appendix}}

\vspace{0.5cm}

\addtocontents{toc}{\protect\setlength{\parskip}{1pt}}
\addcontentsline{toc}{part}{Appendix} 

\etocsettocdepth.toc{none}

\etocignoretoctocdepth
\etocsetnexttocdepth{subsection}
\localtableofcontents

\vspace{0.3cm}

\section{Additional Theoretical Results}
\label{app:theory}

\subsection{Equivalence of Minimum Description Length and Kolmogorov Complexity}
\label{app:equivalenceOfKolmCompAndDescLength}
Here, we briefly show the following Lemma for the reader.
\begin{lemma}
    Up to small constants, the Kolmogorov complexity $K_U$ defined in Eq.~\eqref{eq:KolmComp} and the minimum of the description length $L_\mu$ defined in Eq.~\eqref{eq:descriptionLength} are equivalent if we take $\mathcal{M}$ to be the space of all computable probability distributions over the set of all finite sequences over some alphabet.
\end{lemma}
\begin{proof}
    Given any distribution $\mu$, arithmetic coding allows to encode any string $x$ into a code of length no more than $-\log_2(\mu(x))+c_0$ bits, where $c_0$ is a small constant if we assume sufficient numerical precision for the encoding. Hence we know that for each computable distribution $\mu$, there is a program with length at most $\ell(AC)+\ell(\mu)-\log_2(\mu(x))+c_0 = L_\mu + \ell(AC) + c_0 =: L_\mu + c$ bits that generates $x$, where $\ell(AC)$ is the (usually small) length of the program that performs arithmetic (de)coding on the fixed universal machine $U$ (and we defined $c:=\ell(AC) + c_0$). 
    
    Since $K_U(x)$ is the length of the shortest possible program that generates $x$, it must in particular be shorter than $L_\mu + c$ for all $\mu$. Thus $K_U(x) \le c + \min_{\mu\in \mathcal{M}}L_\mu(x)$. 
    
    Conversely, we can use $p^*(x):=\arg\min_p \{ ~\ell(p)~|~U(p)=x~\}$ to define a distribution $\lambda_x$ by 
    \begin{equation}\begin{split}
            \lambda_x(y) = \begin{cases}
                1,&\text{if }y=U(p^*(x)),\\
                0,&\text{else.}
            \end{cases}
    \end{split}\end{equation}
    Then, up to a small constant $c'$, $\ell(\lambda_x)$ is equal to $\ell(p^*(x))=K_U(x)$, while $-\log_2(\lambda_x(x))=0$. Thus we get $\min_\mu L_\mu(x) \le L_{\lambda_x}(x) = \ell(\lambda_x) = K_U(x) + c'$. 
    
    Thus, for all $x$, up to usually small constants $c$ and $c'$, that only depend on $U$ and not on $x$, $K_U(x)$ and $\min_\mu L_\mu(x)$ are equivalent.
\end{proof}

\subsection{Refined Gaussian Square Loss}
\label{app:refinedSquareLoss}

Here we more carefully re-derive Eq.~\eqref{eq:GaussianSquareLoss} and show theoretically and empirically that the argmin results in the correct standard deviation $\sigma$ if the data generating process is given by a conditional Gaussian probability density and enough samples are provided during training.

As explained around Eq.~\eqref{eq:descriptionLength}, the description length for some dataset $X = \{(x_i, y_i)\}_{i \in \{1,\cdots,n\}}$ given some probabilistic model $\mu$ is (up to a small constant) equal to $ -\sum_i \log_2(\mu(x_i)) + \ell(\mu) $,
where $\ell(\mu)$ is the description length of $\mu$.

Now suppose that the probability $p_{\theta,\sigma}$ is described by the density $p_{\theta,\sigma}$, and assume that $\{p_{\theta,\sigma}\}_{\theta,\sigma}$ is a family of conditional Gaussian densities, that is
\begin{equation}
        p_{\theta,\sigma} (b|a) := \frac{1}{\sqrt{2 \pi \sigma^2}} \exp\left(- \frac{\left(b - f_\theta (a)\right)^2}{2 \sigma^2}\right),
\end{equation}
with $\sigma \in \mathbb{R}$ and $\theta \in \mathbb{R}^n$ and $f_\theta$ the function calculated by the parameters $\theta$ of a neural network.

If we want to encode a datapoint $y_i$, then we need the probability $p_{\theta,\sigma}(y_i)$. 
However, we only have a density and the Lebesgue measure of a single point vanishes. 
This is why, in a continuous space, arithmetic coding is not applicable. However, we can discretize the space, for example using Float32 or Float64 precision. 
Suppose that $A_i$ is the smallest positive area around $y_i$ on which $p_{\theta,\sigma}(y_i|x_i)$ is constant. 
For example, this could be the smallest positive Float32 number. However, bigger floats are more sparse than smaller floats and hence have a bigger $A_i$. Alternatively, one could use a subset of Floats with uniform spacing as is common when sampling pseudorandom numbers uniformly between $0$ and $1$. In that case, all $A_i$ would have equal size. In any case, since $p_{\theta,\sigma}$ is constant on $A_i$, we can easily evaluate the following integral and hence compute $p_{\theta,\sigma}(x_i)$ as follows:
\begin{equation}\begin{split}
        p_{\theta,\sigma}(y_i) &= \int_{A_i} dx~p_{\theta,\sigma}(y_i|x_i)  \mu(y) = A_i ~p_{\theta,\sigma}(y_i|x_i) \mu(y)
\end{split}\end{equation}
If $y$ is sampled uniformly, then $\mu(y)$ would be $1/N$ where $N$ is the cardinality of the sample space. The description length becomes
\begin{equation}\begin{split}
        & \ell(p_{\theta,\sigma}) - \sum_i \log_2(A_i ~p_{\theta,\sigma}(y_i|x_i)/N) \\&= 
        \ell(\theta,\sigma) - \sum_i \log_2(p_{\theta,\sigma}(y_i|x_i)) -\sum_i \log_2(A_i)  + \sum_i \log_2(N)
\end{split}\end{equation}
What is important in this last expression is that the last two sums do not depend on $(\theta,\sigma)$. Hence, when optimizing over those parameters, both terms can be ignored. This shows that the argmin of ``density description length'' is actually equivalent to the argmin of proper description length. 

We can also refine the second term:
\begin{equation}\begin{split}
        - \sum_i \log_2(p_{\theta,\sigma}(y_i|x_i))
        = 
        \sum_i \left[ \frac{1}{2} \log_2(2 \pi \sigma^2 ) + (y_i-f_\theta (x_i))^2 / ( 2 \log(2) \sigma^2)  \right]
\end{split}\end{equation}
That is interestingly almost equal to the mean-squared-error (MSE) loss but it is more refined in the sense that the standard deviation $\sigma$ is taken into account and co-jointly optimized. Of course, if $\sigma$ is considered a constant, then the argmin of the above term is the same as the argmin of the MSE loss. One can therefore truly consider MSE a crude special case of description length minimization, obtained by specializing to Gaussian families, a uniform $\ell$-regularizer and by ignoring the variation of the distribution. 

One can, however, optimize the following more exact loss:
\begin{equation}\begin{split}
        \mathcal{L}_{\text{Gauß}} (\theta, \sigma, X) = \frac{1}{n} \left[\ell(\theta,\sigma) + \sum_i \frac{1}{2} \log_2(2 \pi \sigma^2 ) + \sum_i  \frac{(y_i-f_\theta (x_i))^2}{2 \log(2) \sigma^2}\right].
\end{split}\end{equation}
The normalization factor $1/n  = 1/|X|$ does not change the argmin for any given dataset $X$ but helps to simplify the following arguments.

We provide theoretical guarantees that in the large dataset limit $n\to\infty$, minimizing $\mathcal{L}_{\text{Gauß}}$ recovers the exact function $f_\theta$ as well as noise level $\sigma^2$.

\begin{proposition}
    \label{prop:probGausslossConvergence}
    Let $\theta \in \mathcal{R}^n$ be the parameters of a neural network, from a family of neural networks $\Theta$ satisfying the condition $\max_{\theta \in \Theta} \ell_0(\theta) < \infty$. Let $X = \{(x_i, y_i) | i = 1,\dots,n\}$ where the inputs are sampled from a uniform distribution $x_i \sim U(\text{dom}(f_\theta))$ and the targets are sampled from a distribution $y \sim \mathcal{D} = \mathcal{D}(x_i)$
    where $\mathcal{D}$ satisfies the condition:
    $$\forall x_i \in \text{dom}(f_\theta) : \mathbb{E}(y) = f_\theta (x_i) \text{ and } \text{Var}(y) < \infty$$

    In the limit $n \to \infty$
    $$\vartheta_{\text{opt}}, \varsigma_{\text{opt}} := \arg\min_{\vartheta, \varsigma} \mathcal{L}_{\text{Gauss}}$$

    are such that: 
    $$\varsigma_{\text{opt}}^2 = \sigma^2 \text{ and } f_{\vartheta_{\text{opt}}} = f_\theta$$

\end{proposition}

An example of such a distribution $\mathcal{D}$ is the normal distribution $\mathcal{N}(\mu = f_\theta, \sigma^2)$.
The requirements on $\mathcal{D}$ are much weaker though and apply to any basic setting where we encounter a signal $f_\theta (x_i)$ chained with a noise of bounded variance.

{\bf Proof of Prop.~\ref{prop:probGausslossConvergence}:}

For convenience we write:
\begin{equation}
        \mathcal{L}_{\text{Gauß}} (\theta, \sigma, X)
        = \frac{\ell(\theta, \sigma)}{n} + \mathcal{L}_{\text{simple}} (\theta, \sigma, X)
\end{equation}

We may assume that $\ell(\theta, \sigma)$ is bounded by some constant and thus in the limit $n \to \infty$:
\begin{equation}
    \arg\min_{\vartheta, \varsigma} \mathcal{L}_{\text{Gauß}} (\vartheta, \varsigma, X)
    =     \arg\min_{\vartheta, \varsigma} \mathcal{L}_{\text{simple}} (\vartheta, \varsigma, X)
\end{equation}
It suffices to prove the Proposition for $\mathcal{L}_{\text{simple}}$ rather than $\mathcal{L}_{\text{Gauss}}$.

Further, we use the following helper lemmas (without proof):

\begin{lemma}
    \label{lemma:helper1}
    Let $x \sim \mathcal{D}$,$ \mu = \mathbb{E}(D)$, $\varsigma^2 = \text{Var}(D)$ for some distribution $\mathcal{D} \text{ over } \mathbb{R}$ satisfying the condition:
    $$\mathbb{E}, \text{ the expectation of } \mathcal{D}, \text{ is a bounded linear operator, and } \text{Var}(\mathcal{D}) \equiv \sigma^2 < \infty$$

    Let $m \in \mathbb{R}$, then
    $$\arg\min_m \mathbb{E}[(x - m)^2] = \mu$$
    and
    $$\min_m \mathbb{E}[(x - m)^2] = \varsigma^2.$$
    An example for $\mathcal{D}$ is the normal distribution $\mathcal{N}(\mu, \varsigma^2)$.
\end{lemma}

\begin{lemma}
    \label{lemma:helper2}
    Let $f = f(a) > 0$, $g = g(a, b) > 0$ positive real functions and $b^*(a) := \arg\min_b g(a, b)$,
    then the minimum of $f(a) + g(a, b)$ occurs at $b = b^*(a)$.
    I.e. it holds that: $$\min_{a,b} [f(a) + g(a,b)] = \min_a [f(a) + g(a, b^*(a))]$$
\end{lemma}

The law of large numbers tells us that for large $n$, the empirical average approximates the expectation value; in the limit $n \to \infty$, equality holds. Assuming $n$ is large, we can thus make the approximation:
\[
\mathcal{L}_{\text{simple}}(\vartheta, \varsigma, X) \approx \mathbb{E} \left[ \frac{1}{2} \log_2(2 \pi \varsigma^2) + \frac{(y_i - f_\vartheta (x_i))^2}{2 \log(2) \varsigma^2} \right]
\]

Due to linearity of expectation:
\[
\mathbb{E} \left[ \frac{1}{2} \log_2(2 \pi \varsigma^2) + \frac{(y_i - f_\vartheta (x_i))^2}{2 \log(2) \varsigma^2} \right]
= \frac{1}{2} \log_2(2 \pi \varsigma^2) + \frac{1}{2 \log(2) \varsigma^2} \mathbb{E}\left[(y_i - f_\vartheta (x_i))^2 \right]
\]

Evaluating the $\arg\max$ with respect to $\vartheta$ and $\varsigma$ of this term, we can apply Lem.~\ref{lemma:helper2} to conclude that the optimal $\vartheta$ minimizes
\[
\frac{1}{2 \log(2) \varsigma^2} \mathbb{E} \left[ (y_i - f_\vartheta (x_i))^2 \right]
\]
for a given $\varsigma$.

We remind ourselves that $y_i \sim \mathcal{N}(f_\theta(x_i), \sigma^2)$.

Since, for any $a > 0$, 
\[
\arg\min_x f(x) = \arg\min_x a f(x),
\]
we can apply Lem.~\ref{lemma:helper1} to conclude that $\vartheta_{\text{opt}}$ must be such that $f_{\vartheta_{\text{opt}}}(x_i) = f_\theta (x_i)$ and $(y_i - f_{\vartheta_{\text{opt}}}(x_i))^2 = \sigma^2$.

With this, the second half of the statement is already proven.

To show that $\varsigma_{\text{opt}}^2 = \sigma^2$, we can now simplify:
\[
\arg\min_{\theta, \varsigma} \frac{\log_2(2 \pi \varsigma^2)}{2}  + \frac{1}{2 \log(2) \varsigma^2} \mathbb{E} \left[ (y_i - f_\vartheta(x_i))^2 \right]
= \arg\min_{\varsigma} \frac{\log_2(2 \pi \varsigma^2)}{2}  + \frac{\sigma^2}{2 \log(2) \varsigma^2} 
\]
Solving this problem using classical optimization yields that 
$$f(\varsigma) = \frac{\log_2(2 \pi \varsigma^2)}{2}  + \frac{\sigma^2}{2 \log(2) \varsigma^2} $$
has one local optimum, which is a minimum at $\varsigma^2 = \sigma^2$.

\hfill$\square$

\subsection{Proofs of Results from Sec.~\ref{theory}}
\label{app:theory-proofs}

\subsubsection{Proof of Prop.~\ref{prop:probRewrite}}
\label{proof:prop:probRewrite}

The Bernoulli distribution $\pi_\gamma$ over $\{0,1\}^n$ is given by
\begin{equation}
        \pi_\gamma(z) = \prod_{j=1}^{n} (\gamma_j \delta_{1,z_j}+(1-\gamma_j) \delta_{0,z_j}),
    \label{eq:piz}
\end{equation}
where $\delta_{i,j}$ is $1$ if $i=j$ and $0$ else and $\gamma_j \in [0,1]$.
Hence,
\begin{align*}
        \mathbb{E}_{z\sim \pi_\gamma}[g(z)] &= \sum_{z\in \{0,1\}^n} \pi_\gamma(z) g(z)
        = \sum_{z} \prod_{j=1}^{n} (\gamma_j \delta_{1,z_j}+(1-\gamma_j)\delta_{0,z_j}) g(z) \\
        &=\sum_{z} \prod_{j\ne k}^{n} (\gamma_j \delta_{1,z_j}+(1-\gamma_j)\delta_{0,z_j}) (\gamma_k \delta_{1,z_k}+(1-\gamma_k) \delta_{0,z_k}) g(z)\\
        &= \gamma_k\sum_{\{z|z_k=1\}} \prod_{j\ne k}^{n} (\gamma_j \delta_{1,z_j}+(1-\gamma_j)\delta_{0,z_j}) g(z) \\&\qquad
         + (1-\gamma_k)\sum_{\{z|z_k=0\}} \prod_{j\ne k}^{n} (\gamma_j \delta_{1,z_j}+(1-\gamma_j)\delta_{0,z_j}) g(z).
\end{align*}
Since $\gamma_k$ is bounded and the expression $\mathbb{E}_{z\sim\pi}[g(z)]$ is linear in $\gamma_k$, the expression  must reach its extremal points at the boundaries, i.e. at $\gamma_k=0$ and at $\gamma_k=1$. Since $k$ was arbitrary in the argument above, this must hold for all $\{\pi_i\}_{i\in \{1,\cdots,n\}}$. Therefore, $\min_{\pi}\mathbb{E}_{z\sim \pi}[g(z)]=\min_z g(z)$. 

\hfill$\square$

\subsubsection{On Quadratic Constraint Terms}
\label{sec:squareTerms}

It is important to use a square term $u\cdot(\theta-wz)^2$ for the constrained formulation in Eq.~\eqref{eq:constrained} and Eq.~\eqref{eq:nonlinearProb} because a linear term $u\cdot(\theta-wz)$ would have become $u\cdot(\theta-w\gamma)$, which would not have ensured that $\gamma$ is pushed towards $0$ or $1$. The reason is that one feasible solution would make $\gamma$ vanishingly small (to reduce the loss of the $\ell_0$ term $\sum_i\gamma_i$) and $w$, which is free, equal to $\theta/\gamma$. However, when adopting a quadratic constraint, then we get two terms, namely $u\cdot  (\theta-w\gamma)^2$ and  $u\cdot(w^2\gamma(1-\gamma)) $. If again, we choose $w=\theta/\gamma$ and $\gamma\approx 0$ to bring both $\sum_i \gamma_i$ and the first term towards $0$, then the second term becomes $u\cdot(\theta^2(1-\gamma)/\gamma)\approx u\cdot(\theta^2/\gamma) $ (where division is done componentwise). Since $\gamma$ is small, this becomes very big. As a result, the quadratic formulation evades the problem of the linear constraint that could make the optimization end up in an optimum that we do not want.
Therefore Lem.~\ref{lem:quadratic} is key to make the procedure work well in practice. 

\newpage
\section{Optimization and Algorithmic Details}
\label{app:optimization}

\subsection{ADMM Formulation for \texorpdfstring{$\ell_0$}{L0} Regularization}
\label{app:admm}

\subsubsection{Preliminaries}

The \textit{alternating direction method of multipliers} (ADMM) is an optimization method for constrained optimization problems with stable convergence guarantees for a wide class of problems \cite{boyd2011distributed}. Furthermore, it allows to parallelize operations over multiple processors and compute nodes. In general, it takes the form,
\begin{equation}
        \text{minimize}\quad \sum_i f_i(x_i) + g(z)\qquad \text{ subject to }\quad M_ix_i=Pz+c,~\forall i,
    \label{eq:consensusADMMRescaled}
\end{equation}
with $x_i,z\in\mathbb{R}^n$, $i\in\{1,\cdots,N\}$, and $M_i$ and $P$ arbitrary matrices with appropriate dimensions, and $c$ a constant vector.

An important special case is $M_i=\text{id}=P$, and $c=0$. Then the problem is equivalent to minimization of $\sum_i f_i(x)+g(x)$ but to express it in the form Eq.~\eqref{eq:consensusADMMRescaled} allows for parallel processing, as we will see below. For example $f_i(x)$ could be the $i$-th batch of some model, parameterized by $x$.

We can associate a Lagrangian to this problem, namely
\begin{equation}
        L(\{x_i\}_i,z,\{y_i\}_i) = \sum_i f_i(x_i)+g(z) + y_i(M_ix_i-Pz-c) + g(z),
\end{equation}
where all $y_i$ are vectors, whose elements are Lagrange multipliers. The constrained problem Eq.~\eqref{eq:consensusADMMRescaled} is then equivalent to the following unconstrained minimax problem:
\begin{equation}
        \min_{x_i,z} \max_{y_i} L(\{x_i\}_i,z,\{y_i\}_i).
\end{equation}
At solution points of the problem, the constraints $r_i:=M_ix_i-Pz-c=0$ (also known as the residual) must be satisfied and hence we can augment the Lagrangian with the following term, which is quadratic in the residuals:
\begin{equation*}
        L_\rho(\{x_i\}_i,z,\{y_i\}_i) = \sum_i f_i(x_i)+g(z) + y_i(M_ix_i-Pz-c) + \frac{\rho}{2}||M_ix_i-Pz-c||_2^2 + g(z).
    \label{eq:augmentedLag}
\end{equation*}
We have: $\min_{x_i,z} \max_{y_i} L(\{x_i\}_i,z,\{y_i\}_i)=\min_{x_i,z} \max_{y_i} L_\rho(\{x_i\}_i,z,\{y_i\}_i)$ but at the same time, the quadratic augmentation stabilizes the numerical solution procedure and guarantees convergence for a wider class of functions.

The ADMM procedure now consists of alternatingly minimizing with respect to $x_i$ and to $z$ and maximizing with respect to the Lagrange multipliers $y_i$ (hence the name \textit{alternating direction method of multipliers}). Concretely, the updates are given by the following recursive equations:
\begin{equation}\begin{split}
        x^{k+1}_i &= \argmin_{x_i^k\in\mathbb{R}^n}L_\rho(\{x_i^k\}_i,z,\{y_i\}_i) \\
        &= \argmin_{x_i^k\in\mathbb{R}^n}\left\{\sum_i f_i(x_i^k) + y_i^k(M_ix_i^k-Pz^k-c) + \frac{\rho}{2}||M_ix_i^k-Pz^k-c||_2^2\right\}\\
        z^{k+1} &= \argmin_{z^k\in\mathbb{R}^n}L_\rho(\{x_i^k\}_i,z,\{y_i\}_i) \\
        &= \argmin_{z^k\in\mathbb{R}^n}\left\{g(z^k)+\sum_i\left(y_i^k(M_ix^{k+1}_i-Pz^k-c)+\frac{\rho}{2}||M_ix^{k+1}_i-Pz^k-c||_2^2\right)\right\},\\
        y^{k+1}_i &= y_i^k + \rho(M_ix_i^{k+1}-Pz^{k+1}-c)
    \label{eq:consensusADMMupdateRulesRescaled}
\end{split}\end{equation}
These can be brought into a slightly simpler form by introducing rescaled variables $u_i:=y_i/\rho$.
\begin{equation}\begin{split}
        x^{k+1}_i & 
        = \argmin_{x_i^k\in\mathbb{R}^n}\left\{\sum_i f_i(x_i^k) + \frac{\rho}{2}||M_ix_i^k-Pz^k-c+u^k_i||_2^2\right\}\\
        z^{k+1} & 
        = \argmin_{z^k\in\mathbb{R}^n}\left\{g(z^k)+\sum_i\frac{\rho}{2}||M_ix^{k+1}_i-Pz^k-c+u^k_i||_2^2\right\},\\
        u^{k+1}_i &= u_i^k + M_ix_i^{k+1}-Pz^{k+1}-c
    \label{eq:consensusADMMupdateRulesRescaledForm}
\end{split}\end{equation}
This algorithm is known to converge to a global optimum under the conditions that $f_i$ and $g$ are convex, closed and proper, or equivalently, that their epigraphs are convex sets (cf. \cite{boyd2011distributed}). Though in our applications $f_i$ and $g$ are not convex (but are closed and proper), 
we can still attempt to use the method to reach a local optimum.

\subsubsection{L0 regularized ADMM}

What is to our advantage, when inspecting the equations Eq.~\eqref{eq:consensusADMMupdateRulesRescaledForm} of this method, is that the regularization term $g(z)$ is separated from $f_i(x)$, and only occurs with terms \textit{at most quadratic in} $z$. This means that we can apply our Prop.~\ref{prop:probRewrite}, Eq.~\eqref{eq:l0probterm} and Lem.~\ref{lem:quadratic} to this method in the following way:
\begin{enumerate}
    \item Suppose we have a model $f_\theta$ with parameter $\theta$, that we want to optimize to fit some data $D$ and that we want to distribute our computation with $N$ batches of data $D_i,~i\in\{1,\cdots,N\}$ on $N$ compute nodes, that run in parallel. We can do this by copying the model $z:=\theta$ to $N$ other machines, where we define $x_i:=\text{copy}(\theta)$ and where we also copy the data and model such that $f_i(x_i)=\text{copy}(f_\theta)(D_i)$ and optimize
    \begin{equation}
            \text{minimize}\left\{ \sum_i f_\theta(D_i) + \alpha L_0(\theta) =  \sum_i f_i(x_i) + \alpha L_0(z) \right\},\quad\text{subject to }x_i=z.
    \end{equation}
    We see now that this fits the general scheme of Eq.~\eqref{eq:consensusADMMRescaled} with $M=\text{id}=P$ and $c=0$. For example, we could have $f_i(x_i)=-\log_2(p_\theta(D_i))$, minimizing negative log-likelihood of the data, with $L_0$ constraint as in Eq.~\eqref{eq:descriptionLength} and Eq.~\eqref{eq:l0objective}. However, for now, we keep the more general constraints with $M,P,c$ generic and only specialize to $g(z)=\alpha L_0(z)$.
    \item The big advantage is now that the $L_0(\theta)$ term occurs only within an optimization step, that does not involve $f_\theta(D)$. So we can use Prop.~\ref{prop:probRewrite} to reformulate the $z^{k+1}$-update as follows. First, we introduce two new vector-valued variables $w$ and $Z$ and write $z=wZ$, where $wZ$ is componentwise multiplication, and $w\in\mathbb{R}^n$ while $Z\in \{0,1\}^n$. Then we apply the ideas from Sec.~\ref{sec:probabilistic-reformulations}:
    \begin{equation}\begin{split}
            z^{k+1} & 
            = \argmin_{z^k\in\mathbb{R}^n}\left\{\alpha L_0(z^k)+\sum_i\frac{\rho}{2}||M_ix^{k+1}_i-Pz^k-c+u^k_i||_2^2\right\}\\
            &\overset{\text{(Prop.~\ref{prop:probRewrite})}}{=} \argmin_{w\in\mathbb{R}^n,\pi \in \text{Ber}^n}\mathbb{E}_{Z\sim \pi}\left\{\alpha \sum_j Z_j+\sum_i\frac{\rho}{2}||M_ix^{k+1}_i-c+u^k_i-P(wZ)||_2^2\right\}\\
            &\overset{\text{(Lemma.~\ref{lem:quadratic})}}{=} \argmin_{w\in\mathbb{R}^n,\pi \in [0,1]^n}\bigg\{\alpha \sum_j \pi_j+\sum_i\frac{\rho}{2}||M_ix^{k+1}_i-c+u^k_i-P(w\pi)||_2^2 + \\
            &\qquad\qquad\qquad\qquad\qquad\qquad + \sum_{i=1}^N \frac{\rho}{2} \sum_{p,j} P_{p j}^2 w_j^2\pi_j(1-\pi_j) \bigg\}\\
            &=: \argmin_{w\in\mathbb{R}^n,\pi \in [0,1]^n} L(w,\pi)
    \end{split}\end{equation}
    \item We can attempt to solve for $w$ by computing 
    \begin{equation}\begin{split}
            \frac{\partial L(w,\pi)}{\partial w_\ell} &= -\sum_i \sum_j \rho\bigg( (M_ix^{k+1}_i-c+u^k_i)_j-\sum_q P_{jq}w_q\pi_q\bigg)P_{j\ell}\pi_\ell +\\
            &\qquad +  N\rho \sum_{p} P_{p \ell}^2 w_\ell\pi_\ell(1-\pi_\ell) \\
            &= -\rho \sum_{i,j}(M_ix^{k+1}_i-c+u^k_i)_jP_{j\ell}\pi_\ell
            + N\rho\sum_{j,q}\pi_\ell P_{\ell j}^T P_{jq}\pi_q w_q \\
            &\qquad +  N \rho \sum_{j,q} P_{\ell j}^T P_{j \ell} \pi_\ell(1-\pi_\ell) \delta_{\ell q} w_q\\
            &= -\rho\pi_\ell \sum_{i,j}P_{\ell j}^T(M_ix^{k+1}_i-c+u^k_i)_j
            + \\
            &\qquad +  N \rho \pi_\ell \sum_{j,q}P_{\ell j}^T P_{jq}\bigg( \pi_q+ (1-\pi_q) \delta_{\ell q} \bigg) w_q
    \end{split}\end{equation}
    and setting this to $0$.
    \item In general, this can be non-trivial to solve but for the important special case that $P=\text{id}$, or $P_{jq}=\delta_{jq}$, we obtain
    \begin{equation}
            0 = -\pi_\ell \sum_i(M_ix_i^{k+1}-c+u_i^k)_\ell = N\pi_\ell \hat w_\ell.
    \end{equation}
    This leads to either an arbitrary value for $\hat w_\ell$, if $\pi_\ell=0$, or otherwise to
    \begin{equation}
            \hat w = \frac{1}{N} \sum_i(M_ix_i^{k+1}-c+u_i^k) =: \frac{1}{N}\sum_i r_i^k
    \end{equation}
    This is an average, that does not depend on $\pi$ anymore.
    \item Inserting this back into $L(\hat w,\pi)$, for the case $P_{pj}=\delta_{pj}$, we obtain
    \begin{equation}\begin{split}
            L(\hat w,\pi) &= \bigg\{\alpha \sum_j \pi_j+\sum_i\frac{\rho}{2}||r_i^k-(\hat w \pi)||_2^2 +N \frac{\rho}{2} \sum_j \hat w_j^2\pi_j(1-\pi_j) \bigg\}
    \end{split}\end{equation}
    Since this is linear in $\pi_\ell$ for all $\ell\in\{1,\cdots,n\}$, the only thing we need to do in order to find a minimum, is to check the sign of the gradient with respect to $\pi_\ell$, and move $\pi_\ell$ to the most extreme value in the opposite direction of the sign, admitted within the feasible set. This must be a vertex of the cube $[0,1]^n$. We compute
    \begin{equation}\begin{split}
            \frac{\partial L(\hat w,\pi)}{\partial \pi_\ell} &= \alpha - \sum_i\rho  (r_i^k-(\hat w \pi))_\ell \hat w_\ell + N \frac{\rho}{2} \hat w_\ell^2 ((1-\pi_\ell)-\pi_\ell)\\
            &= \alpha - \rho \hat w_\ell \sum_i (r_i^k)_\ell + N \frac{\rho}{2} \hat w_\ell^2 \\
            &= \alpha - \frac{\rho}{2} N\hat w_\ell^2 
    \end{split}\end{equation}
    
    As a consequence, whenever $\alpha - \frac{\rho}{2}N\hat w_\ell^2 <0$, then we set $\pi_\ell=1$ and if it is $\ge 0$, we set $\pi_\ell=0$.
\end{enumerate}
To summarize the result of all steps above, our optimization step for $z^{k+1}=w^{k+1}\pi^{k+1}$ is:
\begin{equation}\begin{split}
        w^{k+1}_\ell = \frac{1}{N}\sum_{i=1}^N \left(M_ix_i^{k+1} - c + u_i^k\right)_\ell,\qquad \pi^{k+1}_\ell = \begin{cases}
            1,&\text{if }2\alpha<\rho N (w^{k+1}_\ell)^2 ,\\
            0,&\text{else.}
        \end{cases}
    \label{eq:endSol}
\end{split}\end{equation}
This simple exact solution consists of a hard threshold operator that can be carried out in a single step, and can be considered a generalization of hard thresholding in compressive sensing.

The final scheme is thus:
\begin{equation}\begin{split}
        x^{k+1}_i & 
        = \argmin_{x_i^k\in\mathbb{R}^n}\left\{\sum_i f_i(x_i^k) + \frac{\rho}{2}||M_ix_i^k-z^k-c+u^k_i||_2^2\right\}\\
        z^{k+1} & 
        = w^{k+1}\pi^{k+1},\quad\text{as given in Eq.~\eqref{eq:endSol}}\\
        u^{k+1}_i &= u_i^k + M_ix_i^{k+1}-z^{k+1}-c
    \label{eq:finalADMMequations}
\end{split}\end{equation}

To handle the $x$-update step, we can train a possibly nonlinear model via gradient descent until close to convergence and then start alternating, always providing the previous $x^k$ as warm start.

\subsection{Layerwise Pruning}
\label{app:layerwisePruning}

The slight generalization of Lem.~\ref{lem:quadratic} that is needed for Layerwise Pruning reads
\begin{equation}\begin{split}
        &\mathbb{E}_{z\sim\pi_\gamma}\bigg[\sum_{k,i}\big(y_{ki}^{\ell} - \sum_j w_{kj}^{\ell}z_{kj}^{\ell}X_{ji}^{\ell}\big)^2\bigg] 
        \\&
        = \sum_{k,i}\big(y_{ki}^\ell - \sum_j w_{kj}^\ell \gamma_{kj}^\ell X_{ji}^\ell\big)^2 
        + \sum_{k,i,j} (w_{kj}^\ell)^2 \gamma_{kj}^\ell (1-\gamma_{kj}^\ell) (X_{ji}^\ell)^2 .
    \label{em:quadraticLayerwise}
\end{split}\end{equation}
If we now suppose that $X_{ji}^{\ell}$ is the activation of layer $\ell-1$, and that the current weights of layer $\ell$ are $\hat w_{kj} \hat z_{kj}$, then we can set $y_{ki}^{\ell}=\sum_j \hat w_{kj} \hat z_{kj} X_{ji}^{\ell}$ and thereafter minimize Eq.~\eqref{em:quadraticLayerwise} with respect to $w,\gamma$ with gradient descent until we converge to an optimum where all components in the vector $\gamma$ are either $0$ or $1$. Thereafter, we set the new weight matrix equal to $w$. This also works seamlessly with biases. Indeed, we can think of $w$ as a matrix whose last column contains the bias values and add an additional row of ones to $X$ to cover this case with the same theorem and algorithm.

We additionally found that it is advantageous to traverse the network in reverse order when combining this method with Random Gradient Pruning introduced in Sec.~\ref{sec:random-gradient-pruning}: We start with this procedure at the last layer and prune away all unnecessary weights. This then produces spurious weights as described in Sec.~\ref{sec:random-gradient-pruning} and App.~\ref{sec:rgpIllustration}. Those weights can thereafter be eliminated efficiently with Random Gradient Pruning before proceeding to the next-to-last layer and so on. In this way, part of the layerwise information is propagated through the entire network.

\subsection{Threshold Adaptive Mask Determination (TAMADE)}
\label{sec:tamade-algo}

The binary search method discussed in Sec.~\ref{sec:adaptive-mask-determination} can be described with the pseudocode in Alg.~\ref{alg:tamade}. It turned out to be highly effective in pruning the right amount of weights, given a tolerated loss increase. Since this allows to replace a threshold hyperparameter that is highly coupled with other regularization parameters with a tolerance parameter that does not exhibit this coupling (and can be kept constant), this effectively eliminates one hyperparameter search dimension from all our experiments. As this helped us tremendously to increase computational efficiency, we view this simple technique as an important contribution.

\begin{algorithm}[ht]
    \caption{TAMADE}
    \label{alg:tamade}
    \vspace{2mm}
    \textbf{Input}: loss function $L$, neural network $f_\theta$, target $t$, lower bound for threshold $low$, upper bound for threshold $high$, tolerance $tol$, binary search resolution $r$ \\
    \textbf{Output}: Highest threshold that does not increase loss by more than $100\cdot tol$ percent.\\
    \textbf{Remark}: Often $t$ is initialized as $L(f_\theta)$ (the current loss), and $low$ as $0$, and $high$ as the highest absolute value of the weights in $\theta$. A good default value for $r$ is 1e-7.\\
    \begin{algorithmic}[1] 
        \STATE $best = low$
        \STATE $steps = 0$
        \WHILE{$|high - low| > r$}
        \STATE $steps = steps + 1$
        \STATE $mid = (low + high) / 2$ \COMMENT{Typical binary search step}
        \STATE $\theta' = \text{copy}(\theta)$
        \STATE $\theta'[|\theta'| \le mid] = 0$ \COMMENT{Set all weights to $0$ whose absolute value is lower or equal to the current threshold value.}
        \IF{$L(f_{\theta'}) \le (1+tol)\cdot t$}
        \STATE $best = mid$ \COMMENT{If loss is within tolerance, then we have a new best threshold value.}
        \STATE $low = mid$ \COMMENT{However, we do not return this best value but instead continue the loop with the new value $low=mid$ because we might find an even higher threshold that results in a loss within tolerance. This is a variation of standard binary search.}
            \ELSE
            \STATE $high = mid$ \COMMENT{Otherwise, we decrease the highest possible threshold value.}
        \ENDIF
        \ENDWHILE
        \STATE \textbf{return} $best, steps$
    \end{algorithmic}
    \text{\\}
\end{algorithm}

\subsection{Illustration of Random Gradient Pruning}
\label{sec:rgpIllustration}

\begin{figure}[ht]
    \centering
    \includegraphics[width=.4\textwidth]{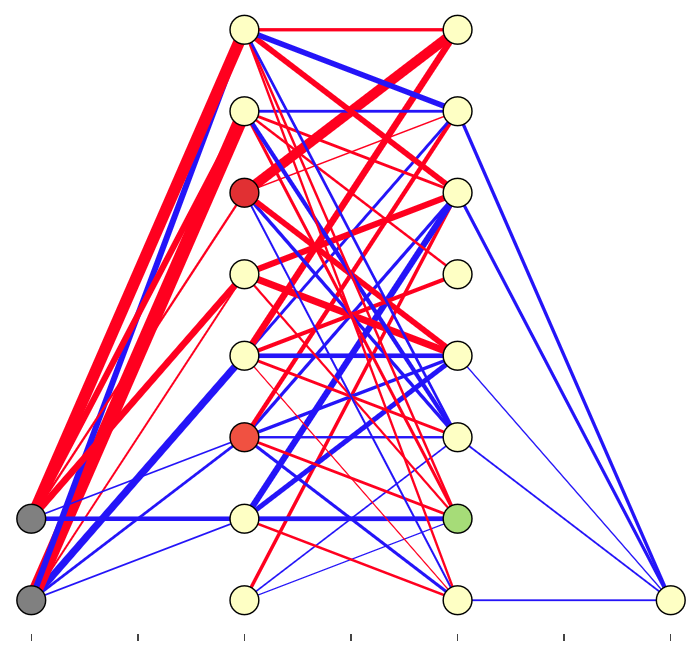}$\qquad\qquad$
    \includegraphics[width=.4\textwidth]{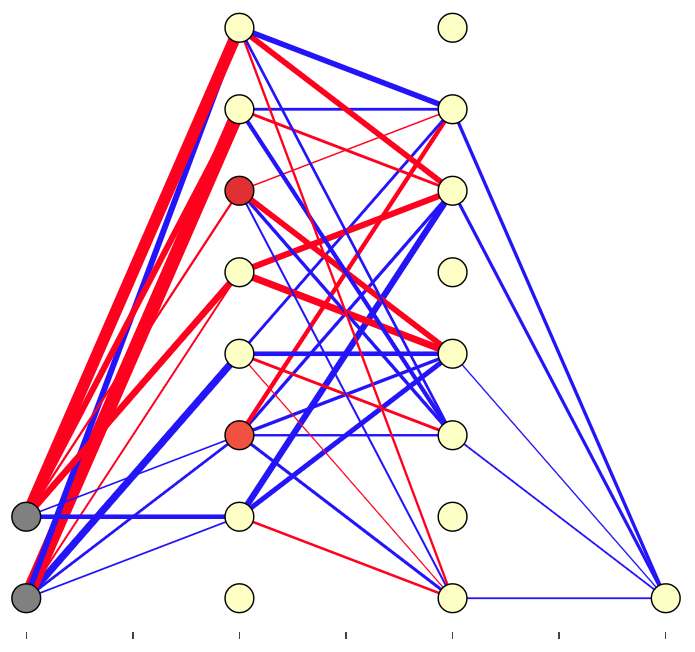}
    \caption{Weights in a simple neural network before (left) and after (right) random gradient pruning. (Line thickness and color indicate connection strength and sign of weights. Circle colors indicate strength of biases.) }
    \label{fig:rgpIllustration}
\end{figure}

To illustrate the discussion in Sec.~\ref{sec:random-gradient-pruning}, Fig.~\ref{fig:rgpIllustration} shows a two-hidden-layer network. On the left side, some method pruned the network that left behind spurious connections that do not contribute to the function that the network computes. For example the first, fourth and seventh neuron in the second to last layer are not connected to the final neuron and the last neuron in the second layer is not receiving any inputs, while its associated bias is zero. Random gradient pruning removes those spurious weights as one can see on the right-hand-side. 

\section{Experimental Methodology}
\label{app:methodology}

\subsection{Experiments on the DRR Method}
\label{app:drr-experiments}

This appendix presents a comprehensive set of experiments designed to systematically address the four primary limitations of the Differentiable Relaxation of $\ell_0$ Regularization (DRR) method:

\begin{enumerate}[nosep=true]
    \item Strong gradients near zero may trap parameters in local minima.
    \item Parameters with large magnitudes but low utility may receive weak gradients and be inefficiently optimized.
    \item The relaxation does not yield exact zeros, necessitating a manual pruning threshold.
    \item The method introduces four hyperparameters ($\alpha$, $\beta$, $\rho$, $\epsilon$), complicating tuning.
\end{enumerate}

Our experiments examine and propose targeted remedies for each of these limitations.

\paragraph{Improving the $\ell_0$ Approximation}

To mitigate issues with gradient magnitude (points 1 and 2), we evaluated alternative smooth approximations to the $\ell_0$ indicator function:

\begin{equation}\begin{split}
  \ell_0(x) &\approx 1 - \exp(-\beta |\theta_i|) \\
  \ell_0(x) &\approx 1 - \exp(-\theta_i^2 / (2\beta^2)) \\
  \ell_0(x) &\approx 1 - \frac{1}{(1 + \theta_i^2 / (3\beta^2))}
\end{split}\end{equation}

The first is the exponential form used in \cite{oliveira2024compression}, while the second and third draw inspiration from the Gaussian and Student-$t$ distributions, respectively. Despite the conceptual appeal of these alternatives, our experiments across all three benchmark datasets found that the original exponential formulation either outperformed or matched the others, offering no consistent benefit to switching.

\paragraph{Adaptive Thresholding with TAMADE}

To address the need for manual pruning (limitation 3), we implemented and evaluated TAMADE (Threshold Adaptive Mask Determination), as described in Sec.~\ref{sec:smooth-reformulations}. TAMADE enables automated, performance-driven threshold selection and reduces the effective hyperparameter dimensionality. Our results demonstrate that TAMADE consistently identifies effective thresholds with minimal computational overhead, delivering high compression rates without compromising test performance. All downstream method comparisons incorporate TAMADE.

\paragraph{Reducing Hyperparameter Burden}

To address limitation 4, we first note that TAMADE reduces the search space by one parameter. We further investigated whether additional simplifications to the $(\alpha, \beta, \rho, \epsilon)$ configuration could maintain performance while easing tuning complexity.

To visualize the trade-off between model compression and predictive performance, we use $\alpha$-Hull plots, which capture the optimal frontiers of accuracy versus sparsity. Based on the $\alpha$-shape method \cite{Edelsbrunner1983ontheshape}, implemented via the GMT library \cite{GMT.jl}, this technique provides holistic comparison by outlining the best achievable trade-offs across thousands of hyperparameter configurations.

While previous work \cite{oliveira2024compression} emphasized the importance of $\ell_2$-regularization in tandem with DRR, our exhaustive grid searches suggest otherwise. We compared pure DRR ($\rho = 0$) against several positive $\rho$ values on three datasets and found that the inclusion of $\ell_2$ offers only marginal benefit, limited to certain compression regimes.

\begin{figure}[t]
\centering
\begin{subfigure}{0.32\textwidth}
    \includegraphics[width=\linewidth]{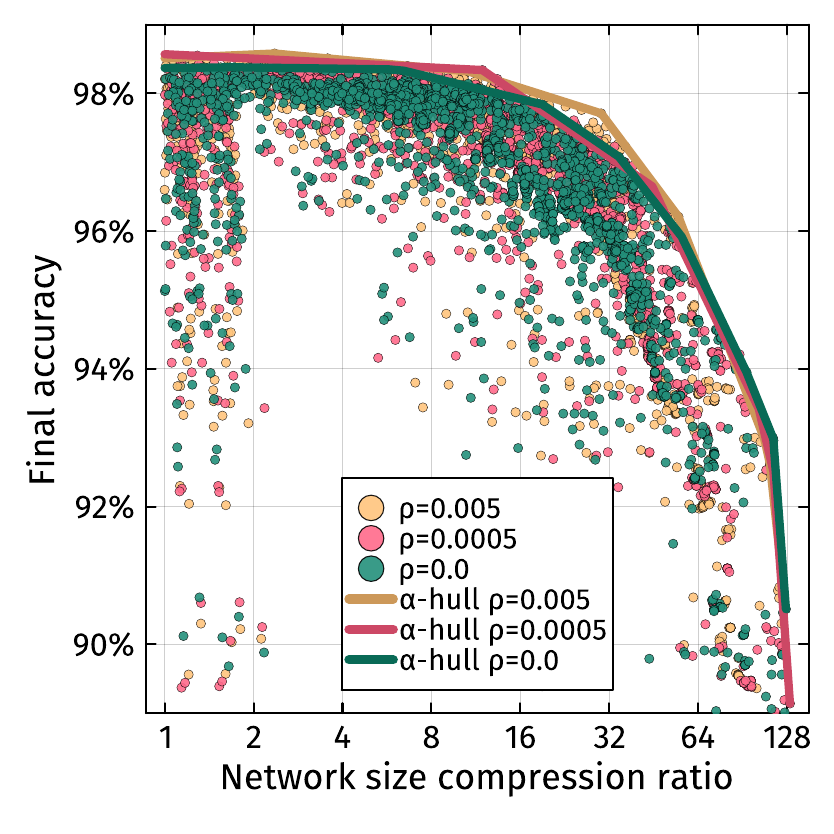}
    \caption{MNIST}
\end{subfigure}
\begin{subfigure}{0.32\textwidth}
    \includegraphics[width=\linewidth]{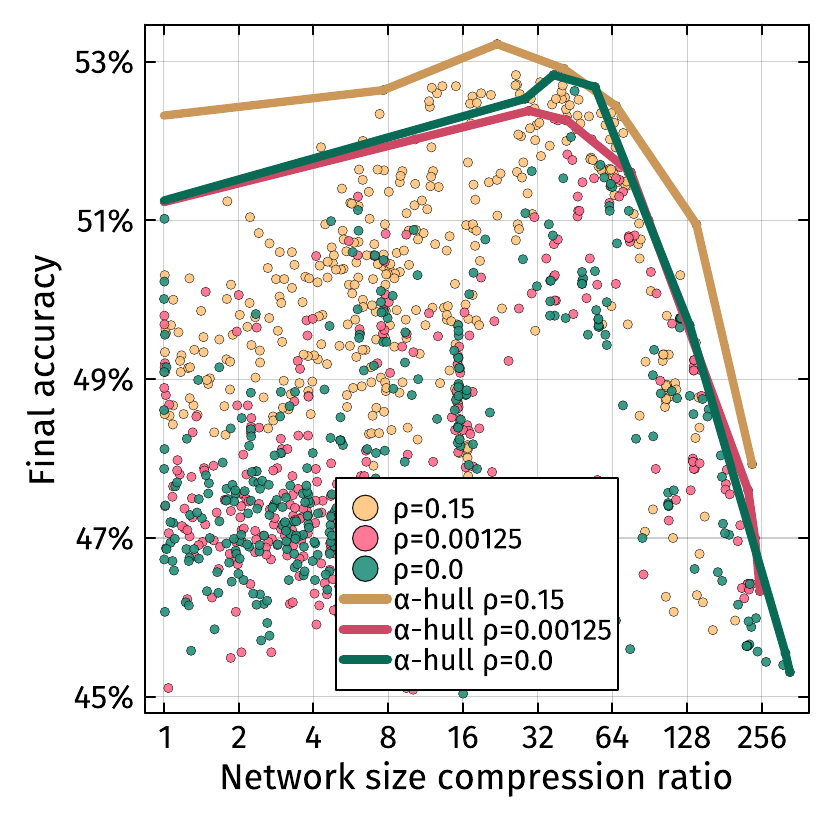}
    \caption{CIFAR-10}
\end{subfigure}
\begin{subfigure}{0.32\textwidth}
    \includegraphics[width=\linewidth]{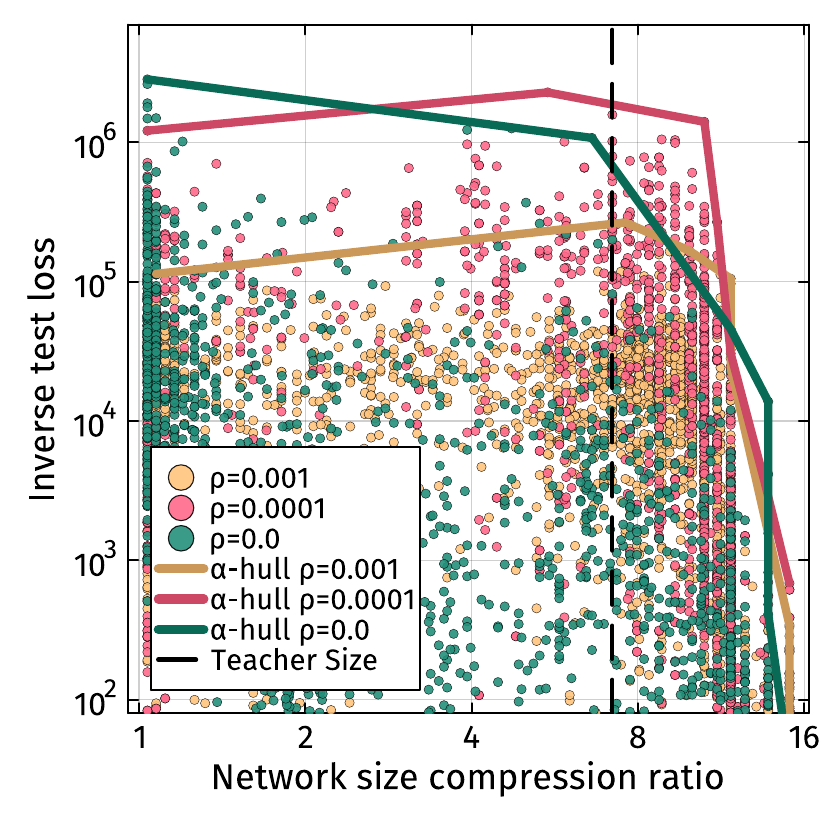}
    \caption{Teacher-Student}
\end{subfigure}
\caption{Effect of $\ell_2$-regularization strength ($\rho$) on performance across datasets, visualized via $\alpha$-Hull plots. Each dot corresponds to a different hyperparameter configuration. Solid lines denote the optimal trade-off frontiers. Compression is measured as the ratio of uncompressed to regularized model size. Color encodes $\rho$: yellow (highest), pink (intermediate), and green ($\rho = 0$). (a) MNIST (LeNet-300-100), (b) CIFAR-10 (MLP with three 512-d hidden layers), and (c) Teacher-Student setting with teacher [2, 5, 5, 1] and student [2, 25, 25, 1]. In (c), inverse test MSE is used (higher is better); the dashed vertical line indicates teacher model size.}
\label{fig:alpha-hull}
\end{figure}

Fig.~\ref{fig:alpha-hull} highlights that for MNIST, the $\alpha$-Hulls for different $\rho$ values are nearly indistinguishable, indicating no benefit from $\ell_2$ regularization. CIFAR-10 and the teacher-student setup show slight gains in some compression rate regimes, but not enough to justify the additional tuning burden.

We also examined whether $\alpha$ and $\beta$ can be reduced to a single hyperparameter. While this simplification holds for MNIST, where fixing one and sweeping the other yields similar performance independent of the fixed value, it breaks down on more complex datasets like CIFAR-10, where their interplay becomes critical.

\paragraph{Dynamic Regularization Schedules}

A further finding relates to training dynamics: gradually introducing DRR during training improves convergence. We tested both fixed schedules that increase $\alpha$ over time and warm-up phases without regularization. Both approaches led to more stable optimization and stronger final results.

This staged strategy balances early-stage learning (favoring flexibility) with late-stage compression (favoring sparsity), and we recommend it as a general practice when applying DRR in new contexts.

\subsection{Hyperparameter Selection and Ablation Studies}
\label{app_ablation_studies}
\label{app:HyperparametersForMethodComparison}

This appendix provides details on the ablation studies conducted to determine optimal hyperparameters for our comparative experiments between DRR, PMMP, and R-L1 methods.

\subsubsection{MNIST and Teacher-Student experiments}
\label{app:mnist-and-ts-ablations}

\paragraph{Methodology}

For each experiment and method, we tested multiple hyperparameter combinations and selected those that yielded the best results. We evaluated performance by plotting accuracy/loss versus regularization strength $\alpha$, with color information representing description length. For teacher-student experiments, we chose parameters that performed best overall across six different combinations of noise-level and training set size.

The experiments included:
\begin{itemize}[nosep]
\item MNIST with LeNet-300-100
\item MNIST with LeNet-5-Caffe
\item CIFAR-10 with VGG16
\item Teacher with Gaussian Loss
\item Teacher with MSE Loss
\end{itemize}

Our teacher is an MLP with dimensions [2, 5, 8, 1] whereas the student has initial dimensionality [2, 25, 25, 1] throughout all our experiments. We either minimize with MSE or Gaussian loss (see Eq.~\ref{eq:GaussianSquareLoss}).

\paragraph{Important Findings}

For the DRR method, we compared two $\beta$ values: 5 and 25. The NORM approach proposed by de Resende Oliveira et al.~\cite{oliveira2024compression}, which scales hyperparameters with the size of the layers, performed worse than regular implementation and was therefore excluded from our main experiments.

For the PMMP method, we tested six combinations of initial probabilities $p$ and multiplier values $u$:
$p \in \{0.0, 0.5, 1.0\}$ and $u \in \{1.0, 10.0, 100.0\}$

For the R-L1 method, we did not use layerwise pruning, and there were no additional hyperparameters to set beyond the regularization strength $\alpha$.

The following table summarizes the optimal hyperparameter values identified for each method across our experimental settings:

\begin{table}[ht]
\centering
\begin{tabular}{lcc}
\toprule
\textbf{Experiment} & \textbf{Best $\beta$ (DRR)} & \textbf{Best $(p_\text{init}, u\text{-multi})$ (PMMP)} \\
\midrule
MNIST MLP & $5$ & $(0.0, 1.0)$ \\
MNIST Caffe & $25$ & $(0.5, 1.0)$ \\
CIFAR VGG & $5$ & $(1.0, 1.0)$ \\
Teacher Gauss & $5$ & $(1.0, 1.0)$ \\
Teacher MSE & $5$ & $(0.0, 1.0)$ \\
\bottomrule
\end{tabular}
\end{table}

For simplicity in our main experiments, we used $\beta=5.0$ for the DRR method across almost all datasets, as this value performed consistently well in most settings.
For simplicity, we divide training into 3 phases: Regularized training, pruning and finetuning \cite{lecun1989optimal}. However, our methods also allow to alternate regularized training and pruning.

For the more computationally demanding experiments (ImageNet and Wikipedia), we conducted additional ablations to determine suitable hyperparameters. These are described in the following subsections.

\subsubsection{Layerwise Pruning comparison} 
\label{app:layerwisePruningAblation}

Our MNIST and Teacher-Student ablation studies revealed that our layerwise pruning method does achieve substantial compression, while keeping the function computed by that layer almost invariant. However, its performance was not as good as the performance of our other methods (R-L1, DRR, PMMP) that take all nonlinearities in the model architecture into account.

Combining layerwise pruning with these fully nonlinear methods in alternating fashion occasionally helped when $\alpha$ was set to low values. However, for higher $\alpha$ values, the combination of layerwise pruning with a nonlinear method often did not perform better than using the nonlinear method alone. Based on these findings, we decided not to use layerwise pruning in our main experiments and benchmarks. 

Nevertheless, if pruning specific layers in a network is desired, the method does deliver excellent results, and can serve the community for this purpose.

\subsubsection{ImageNet}

We trained ResNet50 models on ImageNet-1k using DRR ($\beta=5$), R-L1, and PMMP on single NVIDIA L40S GPUs.

For training we use a batchsize of 128 with a SGD  optimizer (momentum 0.9) and a cosine learning rate schedule with 5 warmup epochs and 90 epochs training time (plus optional fine tuning). 
We employ standard data augmentations, namely label smoothing (0.1), color jitter, random cropping and resizing at runtime. 
Our learning rate is 0.09 which is lower than the standard value of 0.1 employed with a batchsize of 256 and higher than scaling the learning rate linearly with the batchsize. We chose this value by ablation over $\{0.05, 0.09, 0.1\}$, optimizing for Top-1 unregularized validation accuracy.
We employ weigh decay by means on $\ell_2$ regularization with a regularization strength of $\rho=8$e-6 (our $\ell_2$ implementation applies $2 \rho w$ which is double of standard weight decay which means that our strength maps to $\rho \approx 4$e-6 which is lower than the standard value of $\rho=1$e-4). Lowering the strength of weigh decay improved validation accuracy in our setup. The reason why we fall back to explicit $\ell_2$ regularization rather than weigh decay employed by the optimizer is that we want to avoid putting $\ell_2$ pressure on the auxiliary parameters used for the PMMP method. 
Our unregularized baseline come out to $72\%$ top-1 validation accuracy used as reference for all error increases reported.

We performed short-horizon runs (10 epochs) to identify a candidate $\alpha$-range for each method before committing to full 90-epoch sweeps.
For R-L1 we swept $\alpha \in [2\text{e-7}, 4\text{e-6}]$ at 90 epochs plus up to 10 fine tuning epochs after pruning. 
Higher regularization yields stronger compression but lower validation accuracy. For the highest tested $\alpha$ values, training collapsed.

For DRR after short-horizon tests, we swept $\alpha \in [1\text{e-8}, 4\text{e-7}]$ with $\beta=5$ throughout.

Lastly, for PMMP we performed ablations to identify ample initial $u$ values after observing that setting the initial value to 0 inhibits training.
We tested $u \in \{2, 3, 5\}$ and $\alpha \in [1\text{e-6}, 1\text{e-5}]$. 
In this range, the specific value of $u$ is not critical to the outcome, however choosing smaller values were found to result in unstable training. 
We tested for different $\rho$ values but found that the original value of $\rho$=8e-6 works best. 
The relationship between (the tested) hyperparameters and training outcome (validation accuracy and compression rate) is less stable than for the other two methods where we find monotonic relations.
Finally, we selected $\alpha$=5e-6 and u=2.

\subsubsection{Transformer experiments}
\label{app:transformer-parameter-ablations}

We use a transformer decoder architecture with sinusoidal positional encoding and multi-head attention with 25 heads, 10 layers, embedding dimension 1600 and a sequence (context window) length of 512 tokens.

For a batch size of 128, we found through a small series of runs that a learning rate of \texttt{3e-4} leads to comparatively low loss and thereafter used this value for all experiments and all other hyperparameter choices. We use 2000 warmup steps, in which we adopt a linear learning rate increase until we reach \texttt{3e-4} and thereafter adopt a cosine learning rate decay schedule.

For the optimizer, we used AdamW \cite{loshchilov2019decoupledweightdecayregularization} with parameters $(0.9, 0.95)$ throughout.
We do not use weight decay for R-L1 or DRR. However, for PMMP we use weight decay values in proportion to regularization strength $\alpha$, as this seems to help avoid bad local optima.

For DRR, we conducted a small ablation study, in which we varied $\beta$ between $1$ and $10$ and found that, similar to what has been observed in the experiment in Section \ref{app:mnist-and-ts-ablations}, that different $\beta$ values just shift the $\alpha$ values at which similar test losses are observed. This implies the positive result that the choice of $\beta$ is not very essential. We subsequently used $\beta=5$ for all remaining runs, though some of the $\beta=10$ runs were included in the plots to save compute.

\subsection{Convergence Criteria}
\label{app:convergence}

Here we describe in more detail the criterion we used to determine whether a given loss curve is saturated, which is employed in the classifier and teacher student simulations in Sections~\ref{sec:ImageClassifierExperiments} and~\ref{sec:stExperiments}. 

Given an array $L$ that consists of the logged loss values (either training or validation loss, one value for each epoch) and given an even integer that we call smoothing window length $w$, we run a function that performs the following computations every $k$ epochs ($k$ determines how often convergence is checked): If the length $\ell(L)$ is lower than $w$, then the function just returns false. Otherwise, it splits the subarray of the last $w$ values in $L$ into two halves, determines the average of each and then checks if the 2nd average is bigger or equal to the first one, up to the variation in the data in those halves.

The slightly innovative part in this procedure is how we determine the variation in the data. Since the data varies along a (usually decaying loss) curve, which has itself varying y-values, we fit smoothed curves to the two halves, subtract the y-values of the smooth curves from the y-values of the data and then compute the standard deviations of the results. To compute the smoothed curves, we use the \texttt{collocate\_data} function from the \texttt{DiffEqFlux} library \cite{rackauckas2020universal}. The \texttt{collocate\_data} function employs an Epanechnikov kernel to determine the smoothed curve. Fitting this smoothed curve is usually so fast that we did not notice a significant increase in time per epoch, even when $k$ is low (and convergence is checked frequently).

The idea behind this more involved criterion is that this process can take into account trends of the loss curve, even when the loss curve itself might be noisy (and thus does not allow for a simpler procedure in which one triggers saturation / convergence as soon as the validation loss increases). Furthermore, it can be used both for validation loss as well as for train loss saturation / convergence. Throughout the paper, we employ validation loss convergence.

Regarding our experiments with transformer decoders and the Wikipedia dataset, we did not use a convergence criterion to determine when to stop training but rather trained for a predetermined number of epochs, as large language models are typically not trained until convergence \cite{touvron2023llama, deepseekai2024deepSeekV3technicalReport}.
The number of iterations was chosen such that the biggest dataset had 20 times more tokens than our transformer had parameters, and training was performed for one epoch, corresponding to findings in \cite{hoffmannTrainingComputeOptimalLarge}. After that single epoch, we pruned the models (except for unpruned transformers) using TAMADE (cf.~Sec.~\ref{sec:adaptive-mask-determination}) and thereafter the models were finetuned for 0.15 epochs.
For the smaller datasets, we then increased the number of epochs such that the total number of iterations stayed roughly constant while making sure that VRAM memory was used efficiently on our machines. For the medium dataset we used 4 epochs of main training and 1 epoch of finetuning and for the small dataset we used 20 epochs of training and 3 epochs of finetuning.

Further details are provided in the codebase of the supplementary material.

\subsection{Computational Resources}
\label{sec:computationalResources}

In this section, we briefly provide an overview of the computational resources needed and used for the simulations described in Sec.~\ref{experiments}.

For a single run of a classifier, as described in Sec.~\ref{sec:ImageClassifierExperiments}, a conventional personal computer with a decent GPU is sufficient. However, to execute the large number of runs that we conducted, we submitted many jobs in parallel to a cluster of Nvidia A100 GPUs.

The MLP experiments described in Sec.~\ref{sec:stExperiments} were executed on a cluster of Intel Xeon CPUs. Since the models and datasets were small, CPU training was much more efficient than GPU training, especially employing Julia's Lux library \cite{pal2023lux, pal2023efficient}. For a single run, a conventional personal computer is again sufficient.

The transformer experiments described in Sec.~\ref{sec:transformerExperiments} were executed on a cluster of A100 GPUs, making use of PyTorch's Distributed Data Parallel (DDP) library. For a single run, data was distributed across up to 16 GPUs that jointly optimized a transformer. Many such jointly optimized runs were in turn run in parallel to obtain all datapoints.

Not all runs were successful, hence the total number of runs exceeded the number of datapoints provided in this paper.

Tab.~\ref{tab:wallclock-times} shows the computational overhead of our regularization methods. All three approaches (R-L1, DRR, and PMMP) introduce minimal overhead, with runtime often increasing under 10\% compared to vanilla training. 
PMMP shows the highest overhead, while R-L1 and DRR are nearly as efficient as unregularized training. This makes our methods computationally extremely efficient.

\begin{table}[ht]
    \caption{Mean wallclock time in seconds per epoch for different methods across datasets. Experiments conducted on NVIDIA A100 GPUs (1 GPU per run for classifier experiments, 16 GPUs per run for transformer experiments).}
    \label{tab:wallclock-times}
    \centering
    \vspace{0.3cm}
    \begin{small}
        \begin{tabular}{lcccc}
            \toprule
            \textbf{Experiment} & \textbf{Unregularized} & \textbf{R-L1} & \textbf{DRR} & \textbf{PMMP} \\
            \midrule
            MNIST MLP & 0.17(5) & 0.20(7) & 0.20(7) & 0.31(8) \\
            MNIST CAFFE & 0.27(1) & 0.30(2) & 0.31(2) & 0.5(1) \\
            CIFAR VGG-16 & 3.26(1) & 3.33(6) & 3.34(7) & 4.0(5) \\
            ImageNet ResNet-50 & 47(2)$\cdot 10^3$ & 44(2)$\cdot 10^3$ & 43(2)$\cdot 10^3$ & 50(1)$\cdot 10^3$ \\
            Wiki 300 MB & 2757(10) & 2780(20) & 2825(10) & 3340(10) \\
            Wiki 1.23 GB & 13.5(1)$\cdot 10^3$ & 13.4(1)$\cdot 10^3$ & 13.5(3)$\cdot 10^3$ & 14.7(1)$\cdot 10^3$ \\
            Wiki 6.16 GB & 53.4(1)$\cdot 10^3$ & 57.1(2)$\cdot 10^3$ & 58.0(2)$\cdot 10^3$ & 68.4(2)$\cdot 10^3$ \\
            \bottomrule
        \end{tabular}
    \end{small}
\end{table}

\newpage
\section{Additional Experimental Results}
\label{app:additional-experiments}


\subsection{Teacher--Student Experiments: Extended Results}
\label{app:TeacherStudentResultSummary}

This section provides additional context for the experiments described in Sec.~\ref{sec:stExperiments}.
Tab.~\ref{teacher-student-description-length} provides a summary of the best description length obtained by optimizing with a given method in a given setting. 

\begin{table}[ht]
    \caption{Description length (train+test) in bytes for teacher-student models across different training set sizes ($N$) and noise levels. 
    Values show median description length with IQR error in parentheses, where "k" indicates kilobytes. Teacher architecture has layer sizes [2, 5, 8, 1]. The test set size was restricted to match the train set size. UP stands for ``unpruned''.}
    \label{teacher-student-description-length}
    \begin{center}
    \begin{small}
    \begin{tabular}{c|cc|cc|cc|c}
    \toprule
    \textbf{N} & \textbf{PMMP} & \textbf{PMMP} & \textbf{DRR} & \textbf{DRR} & \textbf{R-$\ell_1$} & \textbf{R-$\ell_1$} & \textbf{UP} \\
    & (MSE) & (Gauß) & (MSE) & (Gauß) & (MSE) & (Gauß) & \\
    \midrule
    \multicolumn{8}{l}{\textbf{Low noise ($\sigma^2 = 10^{-5}$)}} \\
    \midrule
    30   & 182(2) & 212(5) & 187(3) & 195(2) & 180(1) & 220(11) & 3111(9) \\
    300  & 1637(24) & 1732(28) & 1623(21) & 1623(17) & 1586(18) & 1611(20) & 4199(14) \\
    2000 & 9.4(2)k & 9.0(1)k & 8.6(1)k & 8.8(2)k & 8.86(8)k & 8.79(9)k & 10.86(5)k \\
    \midrule
    \multicolumn{8}{l}{\textbf{High noise ($\sigma^2 = 0.08$)}} \\
    \midrule
    30   & 182(2) & 197(7) & 187(4) & 195(2) & 180(1) & 209(10) & 3105(9) \\
    300  & 1692(4) & 1722(12) & 1675(5) & 1678(5) & 1666(8) & 1668(7) & 4468(6) \\
    2000 & 10.65(3)k & 10.66(3)k & 10.57(1)k & 10.61(2)k & 10.58(2)k & 10.57(1)k & 13.27(2)k \\
    \bottomrule
    \end{tabular}
    \end{small}
    \end{center}
\end{table}

\begin{figure}[ht]
    \centering
    \begin{subfigure}[t]{0.31\textwidth}
        \includegraphics[width=\linewidth]{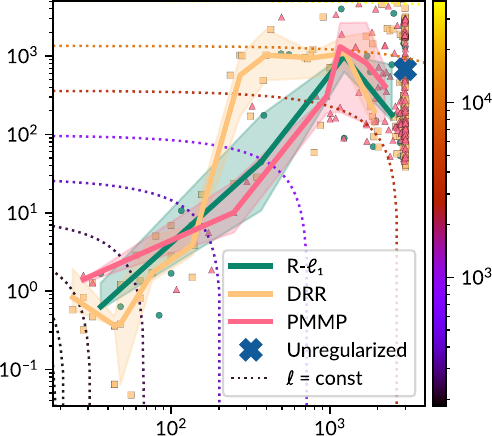}
        \caption{(30, 0.08, Gauß)}
        \label{fig:student-teacher-losses:a}
    \end{subfigure}
    \hfill
    \begin{subfigure}[t]{0.32\textwidth}
        \includegraphics[width=\linewidth]{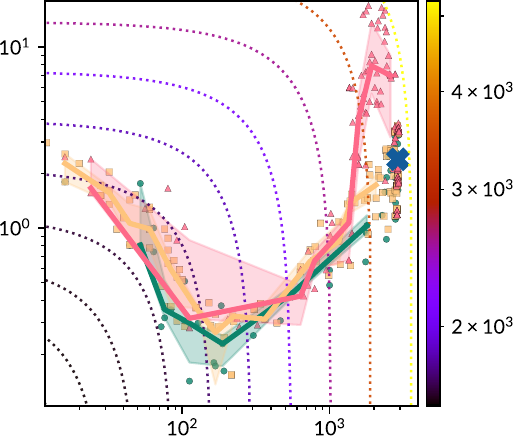}
        \caption{(300, 0.08, Gauß)}
        \label{fig:student-teacher-losses:b2}
    \end{subfigure}
    \hfill
    \begin{subfigure}[t]{0.335\textwidth}
        \includegraphics[width=\linewidth]{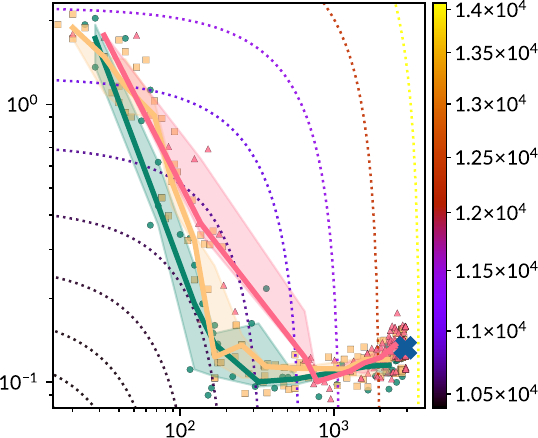}
        \caption{(2000, 0.08, Gauß)}
        \label{fig:student-teacher-losses:c}
    \end{subfigure}
    \\
    \begin{subfigure}[t]{0.3\textwidth}
        \includegraphics[width=\linewidth]{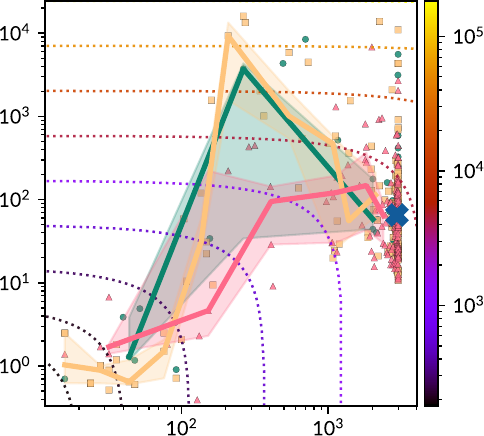} 
        \caption*{(a) (30, 1e-5, Gauß)}
        \label{fig:student-teacher-losses3:a}
    \end{subfigure}
    \hfill
    \begin{subfigure}[t]{0.31\textwidth}
        \includegraphics[width=\linewidth]{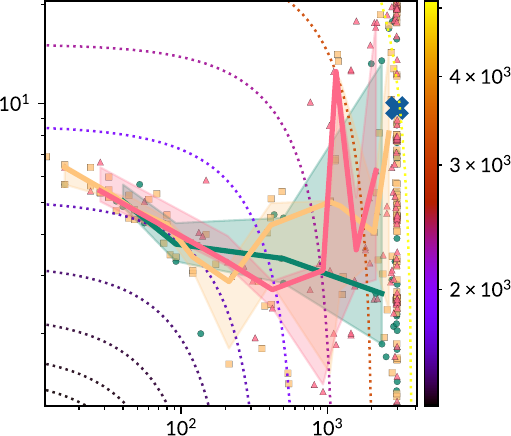} 
        \caption*{(b) (300, 1e-5, Gauß)}
        \label{fig:student-teacher-losses3:b}
    \end{subfigure}
    \hfill
    \begin{subfigure}[t]{0.325\textwidth}
        \includegraphics[width=\linewidth]{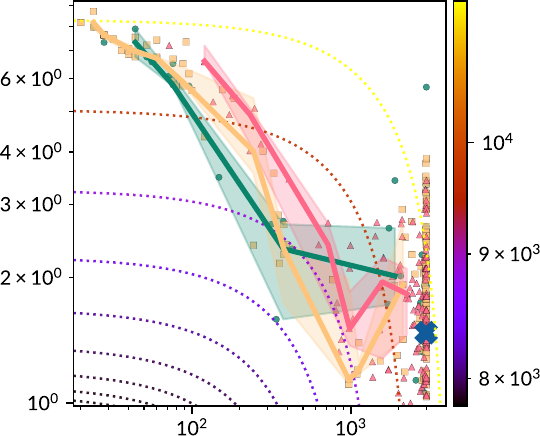} 
        \caption*{(c) (2000, 1e-5, Gauß)}
        \label{fig:student-teacher-losses3:c}
    \end{subfigure}
    \\
    \begin{subfigure}[t]{0.3\textwidth}
        \includegraphics[width=\linewidth]{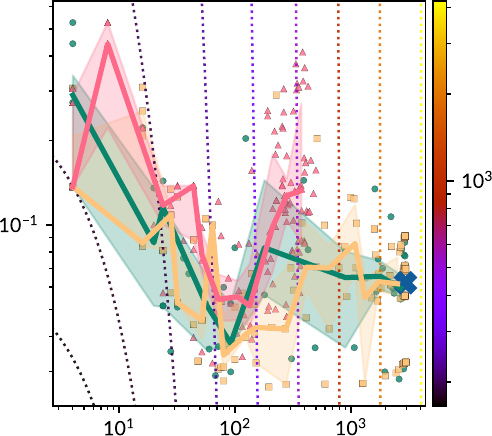} 
        \caption*{(d) (30, 0.08, MSE)}
        \label{fig:student-teacher-losses1:a}
    \end{subfigure}
    \hfill
    \begin{subfigure}[t]{0.31\textwidth}
        \includegraphics[width=\linewidth]{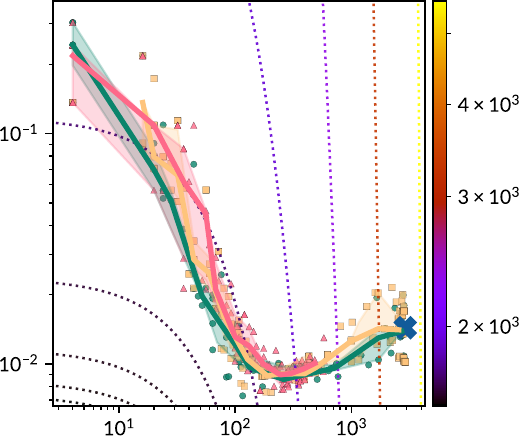} 
        \caption*{(e) (300, 0.08, MSE)}
        \label{fig:student-teacher-losses1:b}
    \end{subfigure}
    \hfill
    \begin{subfigure}[t]{0.325\textwidth}
        \includegraphics[width=\linewidth]{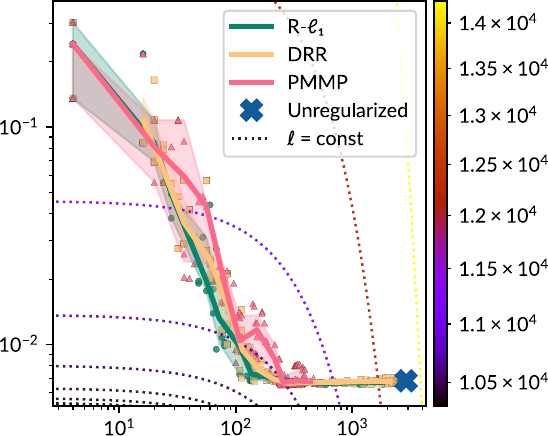} 
        \caption*{(f) (2000, 0.08, MSE)}
        \label{fig:student-teacher-losses1:c}
    \end{subfigure}
    \caption{`Test Loss` vs `Model Byte Size` for different models and dataset sizes in the teacher-student setup. Description length (in bytes) isolines are color coded. Tuples like (2000, 0.08, MSE) denote the dataset size, noise level ($\sigma$) and whether the full Gauß loss Eq.~\eqref{eq:GaussianSquareLoss} or the approximate MSE loss was used. Teacher and student architectures have initial layer sizes [2, 5, 8, 1] and [2, 25, 25, 1], respectively.}
    \label{fig:student-teacher-losses:app}
\end{figure}

Fig.~\ref{fig:student-teacher-losses:app} shows additional plots that exhibit the dependence of test loss on model size (which is in turn determined by regularization strength).
Here, as well as in Fig.~\ref{fig:student-teacher-losses} in the main text and App.~\ref{sec:classifier-benchmark-figures}, the error bars were computed using an adaptive binning approach that groups data points along the x-axis into
bins containing at least 10 points each and spanning at least 0.1 on the log10 scale. For
each bin, the median performance metric was calculated with error tubes indicating the
interquartile range (IQR). This method provides a more robust representation of performance trends compared to simple averaging, especially with the varied outcomes that
result from different hyperparameter combinations and network initializations. One can again observe a U-curve in the center column of the figure and better test accuracy for smaller model sizes for the small dataset in the more accurate Gauß loss scenario, while the MSE loss for the smallest model size is somewhat noisy. For large dataset sizes (right column), the trend is inverted as expected.

\begin{figure}[ht]
    \centering
    \begin{subfigure}[t]{0.3\textwidth}
        \includegraphics[width=\linewidth]{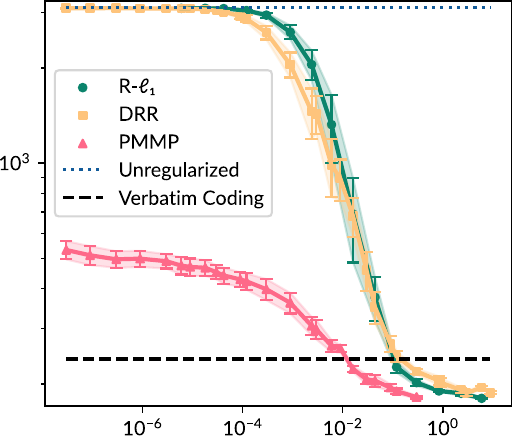} 
        \caption*{(a) (30, 0.08, MSE)}
        \label{fig:student-teacher-description:a}
    \end{subfigure}
    \hfill
    \begin{subfigure}[t]{0.31\textwidth}
        \includegraphics[width=\linewidth]{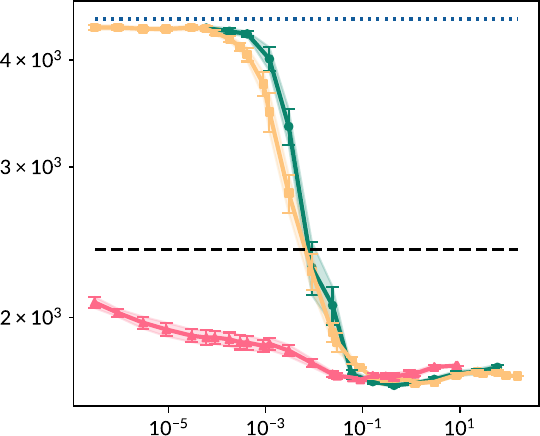} 
        \caption*{(b) (300, 0.08, MSE)}
        \label{fig:student-teacher-description:b}
    \end{subfigure}
    \hfill
    \begin{subfigure}[t]{0.325\textwidth}
        \includegraphics[width=\linewidth]{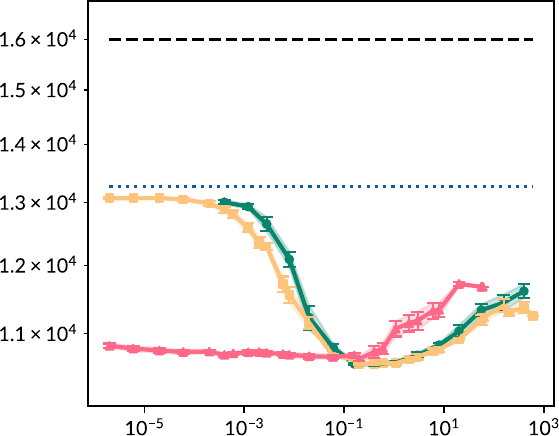} 
        \caption*{(c) (2000, 0.08, MSE)}
        \label{fig:student-teacher-description:c}
    \end{subfigure}
    \caption{`Description Length` vs `$\alpha$` for different teacher student models. Description length (in bytes) isolines are color coded. Each point represents the median result
    from bootstrap sampling across multiple runs for a given regularization strength ($\alpha$), with error
    bars indicating the interquartile range (IQR) of the bootstrap distribution.}
    \label{fig:student-teacher-description-alpha:app}
\end{figure}

Fig.~\ref{fig:student-teacher-description-alpha:app} shows the dependence of description length on $\alpha$. The black line indicates the verbatim coding of the data. One can see that the right amount of regularization depends on the dataset size and is always capable of reducing the description length of the data. As we progress from smaller to larger datasets, we again witness a U-shaped curve pattern emerging (which starts with a left half-U, morphs into a full U and is expected to eventually reverse into a right half-U), interestingly somewhat shifted toward larger dataset sizes compared to the pattern we observe in Fig.~\ref{fig:student-teacher-losses} and~\ref{fig:student-teacher-losses:app}.

\subsection{Transformer Compression Results}
\label{app:TransformerStudies}
\label{sec:additionalTransformerPlots}

\begin{figure*}[ht]
    \centering
    \includegraphics[width=0.32\textwidth]{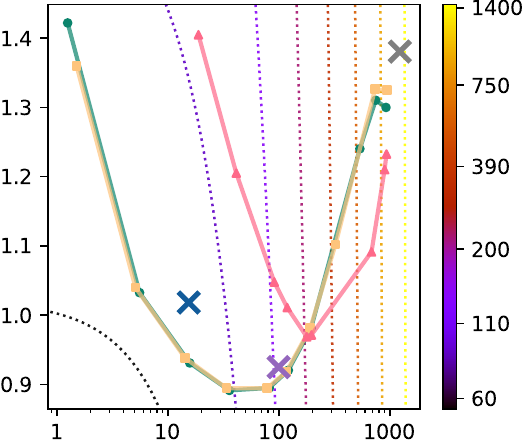}\hfill
    \includegraphics[width=0.32\textwidth]{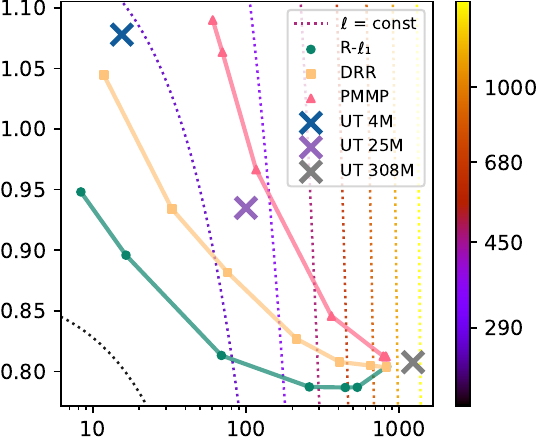}\hfill
    \includegraphics[width=0.32\textwidth]{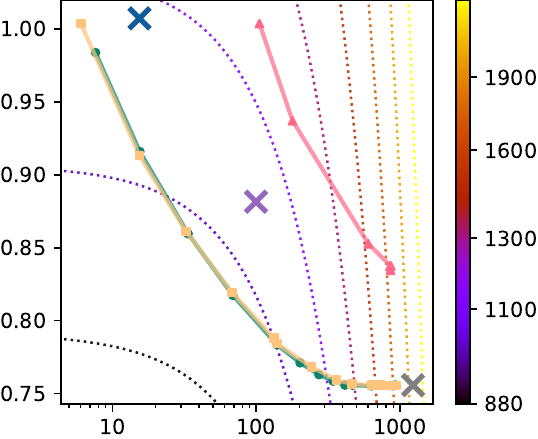}
    \caption{`Mean test loss` vs `model size` (in MB) with color-coded description length (in MB) isolines for the transformer described in Sec.~\ref{sec:transformerExperiments}. From left to right, the 3 panels correspond to dataset sizes of 300MB, 1.2GB and 6.2GB respectively. The legend in the center plot holds for all 3 plots, indicating which solid lines correspond to which regularization method (PMMP, DRR and R-L1, described in Sec.~\ref{sec:probabilistic-reformulations} and \ref{sec:smooth-reformulations}). The crosses correspond to transformers of the indicated model size trained without regularization.}
    \label{fig:transformerTestLossVsModelSize}
\end{figure*}

This subsection provides qualitative analysis supporting the quantitative transformer compression results reported in Sec.~\ref{sec:transformerExperiments} and Tab.~\ref{descLength}.
Fig.~\ref{fig:transformerTestLossVsModelSize} plots mean test loss versus model size for three Wikipedia subsets (300MB, 1.2GB, and 6.2GB), with color-coded isolines indicating total description length (in MB).
Solid curves show the regularized models (PMMP, DRR, R-L1) and crosses indicate unregularized baselines trained at fixed model sizes.

For the 300MB dataset, DRR and R-L1 trace out nearly identical U-shaped curves and consistently outperform unregularized baselines of the same parameter count (albeit the difference is small for the 25M-parameter baseline).
PMMP also exhibits a U-curve but shifted toward larger model sizes.
For the 1.2GB dataset, R-L1 achieves the best performance and retains its U-shape, while DRR does not show a clear U-curve. Both DRR and R-L1 outperform all unregularized baselines, while PMMP performs worse.
For the 6.2GB dataset, DRR and R-L1 are again nearly identical.

A notable effect is that smaller unregularized models (crosses) underperform larger models trained with regularization, especially in the plots corresponding to the bigger dataset, even when both yield the same final parameter count.
This suggests that description-length regularization allows training to select locally optimal parameter subsets from an large parameter space, whereas fixing a reduced architecture \emph{a priori} restricts optimization to a potentially suboptimal subspace.
This effect could be explained by the observation that larger models tend to exhibit more connected loss landscapes \cite{draxler2019essentiallybarriersneuralnetwork}.

Although our transformer (308M parameters, $\approx$1.2GB) is smaller than frontier models, the dataset-to-model size ratios span from data-scarse regimes to the 20-tokens-per-parameter ratio proposed in \cite{hoffmannTrainingComputeOptimalLarge} and matching the regime of models such as LLaMA \cite{touvron2023llama} and DeepSeek-V3 \cite{deepseekai2024deepSeekV3technicalReport}.

\begin{figure*}[ht]
    \centering
    \includegraphics[width=0.32\textwidth]{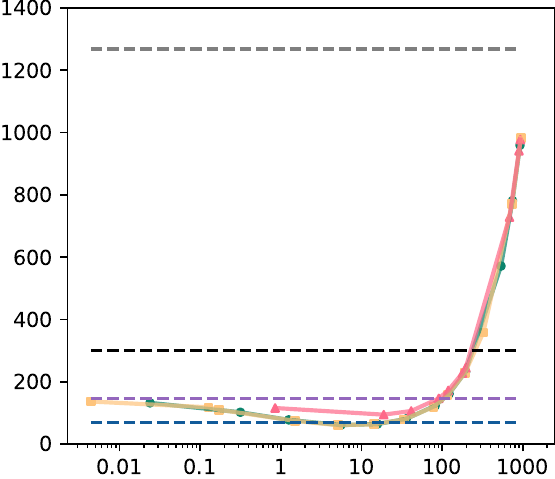}\hfill
    \includegraphics[width=0.32\textwidth]{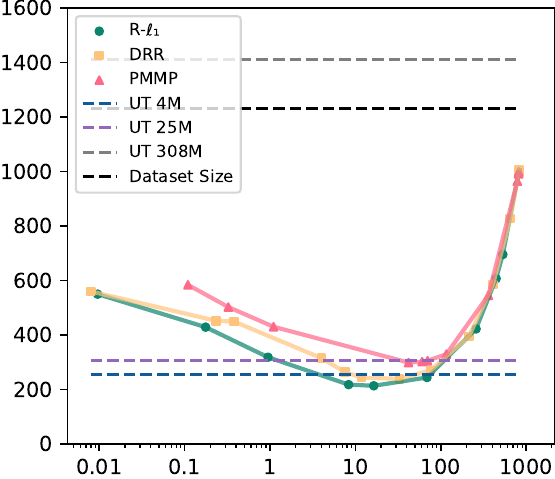}\hfill
    \includegraphics[width=0.32\textwidth]{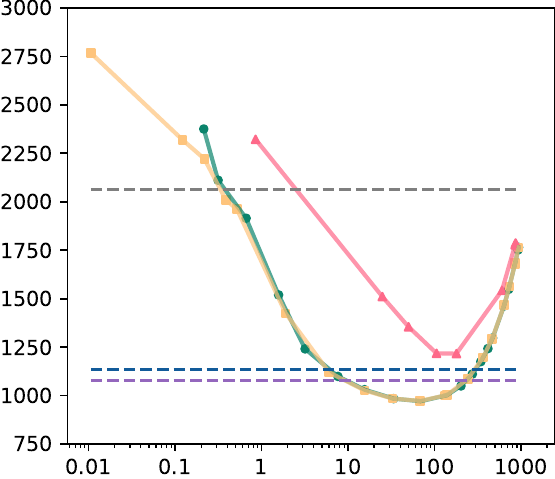}
    \caption{`Description length' vs `model size' (both in MB). From left to right, the 3 panels correspond to dataset sizes of 300MB, 1.2GB and 6.2GB respectively. The legend in the center plot holds for all 3 plots, indicating which solid lines correspond to which regularization method (PMMP, DRR and R-L1, described in Sec.~\ref{sec:probabilistic-reformulations} and \ref{sec:smooth-reformulations}). The horizontal lines correspond to transformers of the indicated model size trained without regularization.}
    \label{fig:transformerDescLengthVsModelSize}
\end{figure*}

Finally, we present description length vs.~model size plots in Fig.~\ref{fig:transformerDescLengthVsModelSize}. As before, the description length is the sum of the model byte size and the coding length of the dataset when encoded using arithmetic code under the probability distribution of the model (offline compression scheme).
Compared to Fig.~\ref{fig:transformerTestLossVsModelSize}, we find a reversed trend:
We find the most pronounced U-shaped curve for the largest dataset while the curve flattens for the two smaller datasets. 
When models become big, their parameter count increases description length, whereas models that are too small they cannot achieve sufficiently small prediction error to yield small arithmetic coding lengths. This explains the U-curves we see in these panels.
For the smallest and middle dataset sizes in the left and middle plot panels, only small models can achieve substantial compression because the model size dominates over the coding length of the dataset. As a result, the left side of the U-shaped curve becomes flatter for smaller datasets.

Similar to the plots in the teacher-student setting, these results in combination with the plots in Fig.~\ref{fig:transformerTestLossVsModelSize} show that optimal test loss is achieved for model sizes slightly large than what is necessary for optimal combined description length minimization. This interesting phenomenon is to our knowledge not yet completely explained by algorithmic information theory and might be the subject of further investigations in the future. However, even without such investigations, it provides a guideline for the optimal choice of model size during training.

\subsection{MNIST Pruning Benchmark}

\label{sec:classifier-benchmark-tables}

We summarize additional benchmarks in Tab.~\ref{tab:mnist-lenet300}. DRR outperforms all previous methods while RL1 delivers comparable but slightly weaker results. PMMP performs less strongly, but on-par with other probabilistic methods such as \cite{louizos2017learning}.

\begin{table}[ht]
    \caption{Pruning on MNIST for two architectures:
    LeNet-300-100 (left EI/CR columns) and LeNet-5-Caffe (right EI/CR columns).
    (EI = Error Increase, CR = Compression Rate).}
    \label{tab:mnist-lenet300}
    \centering
    \begin{small}
        \begin{tabular}{p{2.3cm}cc|cc}
            \toprule
            \textbf{Method} & \textbf{EI} & \textbf{CR} & \textbf{EI} & \textbf{CR} \\
            \midrule
            Autoprune \cite{xiao2019autoprune} & - & 80 & - & 310 \\
            CPA \cite{phan2020pruning} & -0.04 & 85 & -0.02 & 317 \\
            DNS \cite{guo2016DNS} & -0.29 & 56 & 0.00 & 111 \\
            DRR-O \cite{oliveira2024compression} & 0.16 & 90 & 0.16 & 200 \\
            GSM \cite{ding2019globalGSM} & 0.01 & 60 & 0.15 & 300 \\
            $L_0$-ADMM \cite{zhang2018L0ADMM} & 0.00 & 23 & 0.00 & 71 \\
            $L_0$-ARM \cite{li2019L0arm} & - & 14 & - & 196 \\
            $L_0$-HC \cite{louizos2017learning} & - & 10 & - & 93 \\
            $L_0$-$L_2$-LC \cite{idelbayev2022exploringL0L2LC} & - & 50 & - & 100 \\
            SNIP \cite{lee2018snip} & 0.70 & 50 & 0.20 & 100 \\
            SparseVD \cite{molchanov2017variationalSparseVD} & 0.28 & 68 & -0.05 & 280 \\
            SWS \cite{ullrich2017soft} & 0.05 & 23 & 0.09 & 200 \\
            \midrule
            DRR (ours) & 0.16 & \textbf{92} & 0.16 & \textbf{342} \\
            R-L1 (ours) & 0.23 & 68 & 0.08 & 246 \\
            PMMP (ours) & 0.19 & 11 & 0.17 & 77 \\
            \bottomrule
        \end{tabular}
    \end{small}
\end{table}


\subsection{Additional Classifier Benchmark Figures}

\label{sec:classifier-benchmark-figures}

\begin{figure}[ht]
    \centering
    \begin{subfigure}[t]{0.3\textwidth}
        \includegraphics[width=\linewidth]{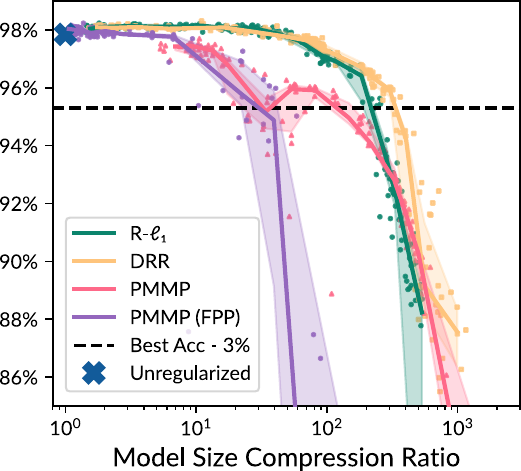}
        \caption*{(a) MNIST with LeNet-300-100}
    \end{subfigure}
    \hfill
    \begin{subfigure}[t]{0.31\textwidth}
        \includegraphics[width=\linewidth]{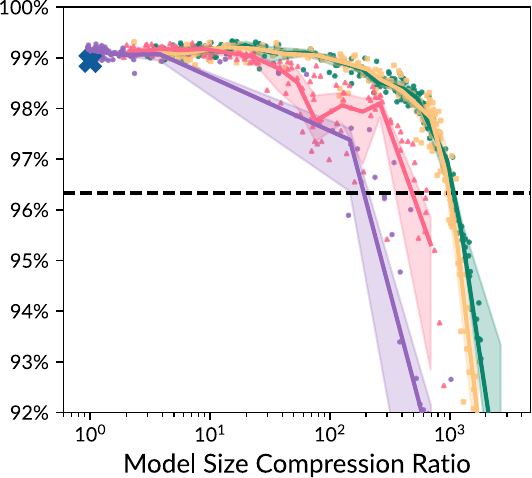}
        \caption*{(b) MNIST with LeNet-5-Caffe}
        \label{fig:caffeAcc}
    \end{subfigure}
    \hfill
    \begin{subfigure}[t]{0.3\textwidth}
        \includegraphics[width=\linewidth]{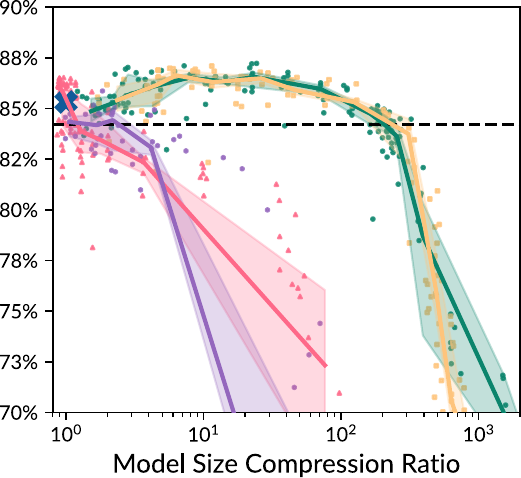}
        \caption*{(c) CIFAR with VGG-16-512}
    \end{subfigure}
    \caption{`Accuracy` vs `Model Size Compression Rate` for different datasets and models.}
    \label{fig:classifierAccuracies}
\end{figure}

To provide additional context for the results described in Sec.~\ref{sec:ImageClassifierExperiments}, we present Accuracy-vs-Compression Rate plots in Fig.~\ref{fig:classifierAccuracies}. As explained at the end of Sec.~\ref{sec:ImageClassifierExperiments}, in addition to R-L1, DRR and PMMP, we benchmarked the performance of \textit{fictitious play pruning} (FPP), which applies the fictitious play \cite{brown1949some,brown1951iterative} optimization strategy to minimize the PMMP objective specified in Corollary \ref{cor:probNonlinear}. 

In fictitious play (in our case adapted to alternating stochastic gradient descent ascent (SGDA), or in fact to alternating ADAM optimizer updates), we alternatingly update two groups of parameters (in our case $u$ and the other parameters) using the parameters of the previous time step of the own group (as in normal SGDA) and (in contrast to SGDA) the \textit{historical averages} of the other parameters in the ``opponent's'' group. This can lead to convergence to a Nash equilibrium when greedy methods fail \cite{brown1949some,brown1951iterative,lee2020efficientexplorationstatemarginal}.

The error bars in those figures were computed using an adaptive binning approach that groups data points along the $x$-axis into bins containing at least 10 points each and spanning at least 0.1 on the log10 scale. For
each bin, the median performance metric was calculated with error tubes indicating the
interquartile range (IQR). This method provides a more robust representation of performance trends compared to simple averaging, especially with the varied outcomes that
result from different hyperparameter combinations and network initializations.

One can observe that DRR and R-L1 generally perform similarly, while PMMP has somewhat lower performance. FPP performs similar to or worse than PMMP. Note that the x-axis is log-scaled and that some of the models achieve very high compression rates before their accuracy falls below the -3\% mark. In the case of CIFAR, one can clearly see how moderate regularization robustly improves test accuracy, confirming our main point that compression can increase generalization capability.

\newpage
\section{Description Length Analysis for LLaMA}
\label{app:llama_analysis}

This appendix provides the methodology and calculations for the LLaMA description length analysis, comparing fixed-model and online learning approaches.

\paragraph{Setup and Methodology}

We analyze the 65B parameter LLaMA model from \cite{touvron2023llama}, trained on approximately 1400B tokens, comparing two approaches to model description length:
\begin{enumerate}[nosep]
    \item Fixed-model approach: Train the model to convergence, then encode data using the final model
    \item Online learning approach: Encode data sequentially as training progresses \cite{bornschein2023sequential,JackRaeVideo}
\end{enumerate}

\begin{figure}[ht]
    \centering
    \begin{subfigure}{0.48\textwidth}
        \includegraphics[width=\textwidth]{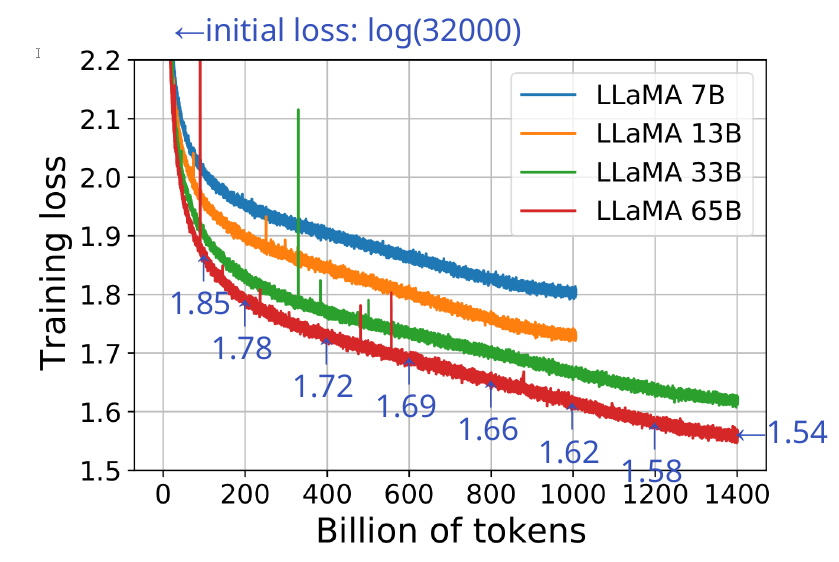} 
        \caption{LLaMA training loss curve from \cite{touvron2023llama}}
        \label{fig:llama-loss}
    \end{subfigure}
    \hfill
    \begin{subfigure}{0.48\textwidth}
        \includegraphics[width=\textwidth]{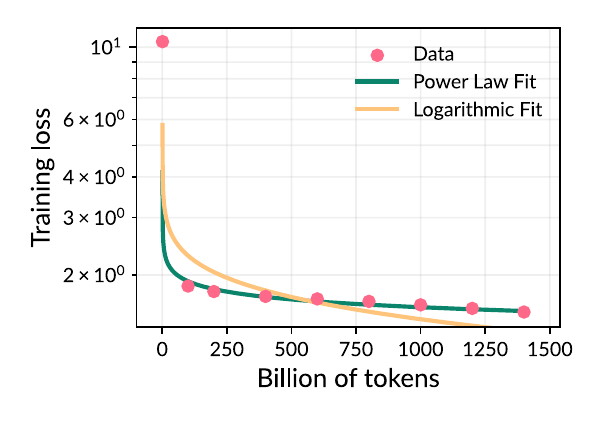} 
        \caption{Power-law curve fit to LLaMA loss data}
        \label{fig:llama-fit}
    \end{subfigure}
    \caption{Loss curves for LLaMA models. Left: Original training curve from \cite{touvron2023llama} showing cross-entropy loss vs. tokens processed, with manually penciled-in loss values (in blue) extracted for analysis. Right: Power-law fit to these extracted 65B model data points.}
    \label{fig:llama-analysis-figures}
\end{figure}

For our analysis, we extracted approximate loss values from the published learning curves. For the initial loss at one token, we used $\log(32000) \approx 10.4$, corresponding to the cross-entropy loss with LLaMA's 32k token vocabulary (assuming uniform distribution). The loss curve was best approximated by the power-law function:

\begin{equation}
y = 2.7553 \cdot (x - 1.0)^{-0.0793}
\end{equation}

\paragraph{Online Learning Approach}

The total description length under the online approach is given by the integral under the loss curve:

\begin{equation}
\int_0^{1400} 2.7553 \cdot (x - 1.0)^{-0.0793} \, \mathrm{d}x = 6349.41
\end{equation}

Converting from base $e$ to storage size (in bytes), this yields approximately 550 GB for the online approach.

\paragraph{Fixed-Model Approach}

For the fixed-model approach, we use the final loss value (approximately 1.54 at 1400B tokens):
\begin{itemize}[nosep]
    \item Data encoding: $1.54 \cdot 1400 \cdot 10^9 = 2156 \cdot 10^9$ in base $e$ (converted to 187 GB in base $2$).
    \item Model size: 131 GB (65B float16 parameters)
    \item Total: 318 GB
\end{itemize}

\paragraph{Results and Discussion}

Comparing the two approaches:
\begin{itemize}[nosep]
    \item Online learning approach: 550 GB
    \item Fixed-model approach: 318 GB (187 GB data + 131 GB model)
\end{itemize}

The fixed-model approach yields a description length approximately 42\% smaller than the online learning approach. This suggests that for current state-of-the-art models, parameter count does impact the total description length meaningfully, supporting the potential value of parameter-reducing techniques.

\newpage
\section{Impact Statement}
\label{app:societalImpact}

The work presented in this article facilitates the compression of data and potentially enables to train models with higher sample-efficiency. It can be used in all machine learning applications where neural models are optimized and sample-efficiency is a concern. Since the work is of a foundational nature, it does not have any direct positive or negative societal impacts. 

\section{Contributions}
\label{app:contributions}

The research presented in this paper was conducted in close collaboration of the authors.
Here we list our contributions, indicating the individual who made the primary contribution to each component.

\begin{center}
\begin{tabular}{p{0.62\textwidth} p{0.3\textwidth}}
\toprule
\textbf{Component} & \textbf{Primary Contributor} \\
\midrule
Conceptualization of research idea and writing of first version & Lukas Barth \\
\addlinespace[4pt]
Probabilistic reformulation of $\ell_0$-regularization including development of PMMP method and ADMM equations & Lukas Barth \\
\addlinespace[4pt]
Development of Gaussian Loss for MDL learning & Lukas Barth \\
\addlinespace[4pt]
Convergence guarantees for Gaussian Loss & Paulo von Petersenn \\
\addlinespace[4pt]
Development of the complementary methods Layerwise Pruning, Random Gradient Pruning, and TAMADE & Lukas Barth \\
\addlinespace[4pt]
Ablations including the improvement of the DRR method & Paulo von Petersenn \\
\addlinespace[4pt]
Statistical analysis and data visualization of simulation results & Paulo von Petersenn \\
\addlinespace[4pt]
Comparative experiments on different $\ell_0$-regularization methods with teacher-student setup, classifiers, and transformers & Equal Contribution \\
\bottomrule
\end{tabular}
\end{center}

\noindent Nevertheless, we discussed each component with each other and consider the entire piece of work as something to which we equally contributed.

\end{document}